\documentclass{article}

\usepackage{microtype}
\usepackage{graphicx}
\usepackage{subfigure}
\usepackage{booktabs} % for professional tables
\usepackage{listings}
\usepackage{amsmath}
\usepackage[ruled,vlined]{algorithm2e}
\usepackage{cuted}
\usepackage{float}
\usepackage{svg}

\usepackage{amssymb}
\usepackage{xcolor}
\usepackage{hyperref}

\makeatletter
\providecommand{\Notice@String}{} % 如果未定义则定义为空
\makeatother
\usepackage[accepted]{wwjjhh2025}
\makeatletter
\providecommand{\wwjjhh@appearing}{} % 如果没定义就当成空
\providecommand{\Notice@String}{}   % 如果没定义就当成空
\makeatother
\wwjjhhtitlerunning{Tree Training: Accelerating Agentic LLMs Training via Shared Prefix Reuse}

\begin{document}

\twocolumn[
\wwjjhhtitle{Tree Training: Accelerating Agentic LLMs Training via Shared Prefix Reuse}

%Trie-packing Training: Packing Tree-based Trajectory in Agentic LLM Training
%Tree Training: Reuse Shared Trajectories for Agentic LLMs
%Tree Training: Reuse Shared Trajectories Efficiently for Agentic LLMs

% unless you've provided the [accepted] option to the wwjjhh2025


\wwjjhhsetsymbol{equal}{*}
\wwjjhhsetsymbol{cor}{\textdagger}

\begin{wwjjhhauthorlist}
\wwjjhhauthor{Jinghui Wang}{equal,cor,1}
\wwjjhhauthor{Shaojie Wang}{equal,1}
\wwjjhhauthor{Yinghan Cui}{equal,1}
\wwjjhhauthor{Xuxing Chen}{equal,1}
\wwjjhhauthor{Chao Wang}{equal,1}
\wwjjhhauthor{Liang Huang}{1}
\wwjjhhauthor{Xiaojiang Zhang}{1}
\wwjjhhauthor{Junyi Peng}{1}
\wwjjhhauthor{Li Wan}{1}
\wwjjhhauthor{Haotian Zhang}{1}
\wwjjhhauthor{Bin Chen}{1}
\end{wwjjhhauthorlist}

% 用数字编号定义单位
\wwjjhhaffiliation{1}{Kwai Inc}

\wwjjhhcorrespondingauthor{Jinghui Wang}{wangjinghui05@kuaishou.com}

\wwjjhhkeywords{Machine Learning, wwjjhh}

\vskip 0.3in

\begin{abstract}
Agentic large language model (LLM) training often involves multi-turn interaction trajectories that branch into multiple execution paths due to concurrent tool use, think-mode, sub-agent, context management and other runtime designs. As a result, the tokens produced by a single task naturally form a tree-structured token trajectory with shared prefixes, rather than a linear sequence. Existing training pipelines linearize such trajectories and treat each branch independently, leading to substantial redundant computation in both forward and backward passes. We derive that averaging the loss over all branches independently is algebraically identical to a per-token weighted loss, where each token's weight equals the fraction of branches passing through it. The problem therefore reduces to computing the log-probability of every token in the prefix tree exactly once, with no repeated computation across shared prefixes: we propose DFS serialization of the tree, which visits every token exactly once, and adapt full-attention and SSM layers to ensure the resulting log-probabilities match independent per-branch calculation exactly. In practice, a single trajectory tree can be too large to fit in GPU memory; we therefore propose \textbf{Redundancy-Free Tree Partitioning}, which handles memory-constrained settings with zero redundant computation and peak memory bounded by a single root-to-leaf path. Together, these contributions form \textbf{Tree Training}, an efficient framework for training LLMs on tree-structured trajectories, achieving up to \textbf{6.2×} end-to-end training speedup on dense and MoE models for both supervised fine-tuning and reinforcement learning.
\end{abstract}
]


% The \wwjjhhEqualContribution command is standard text for equal contribution.

\printAffiliationsAndNotice{\wwjjhhEqualContribution} % otherwise use the standard text.
% \wwjjhhEqualContribution

\begin{figure*}[t]
  \centering
  \includegraphics[width=0.98\linewidth]{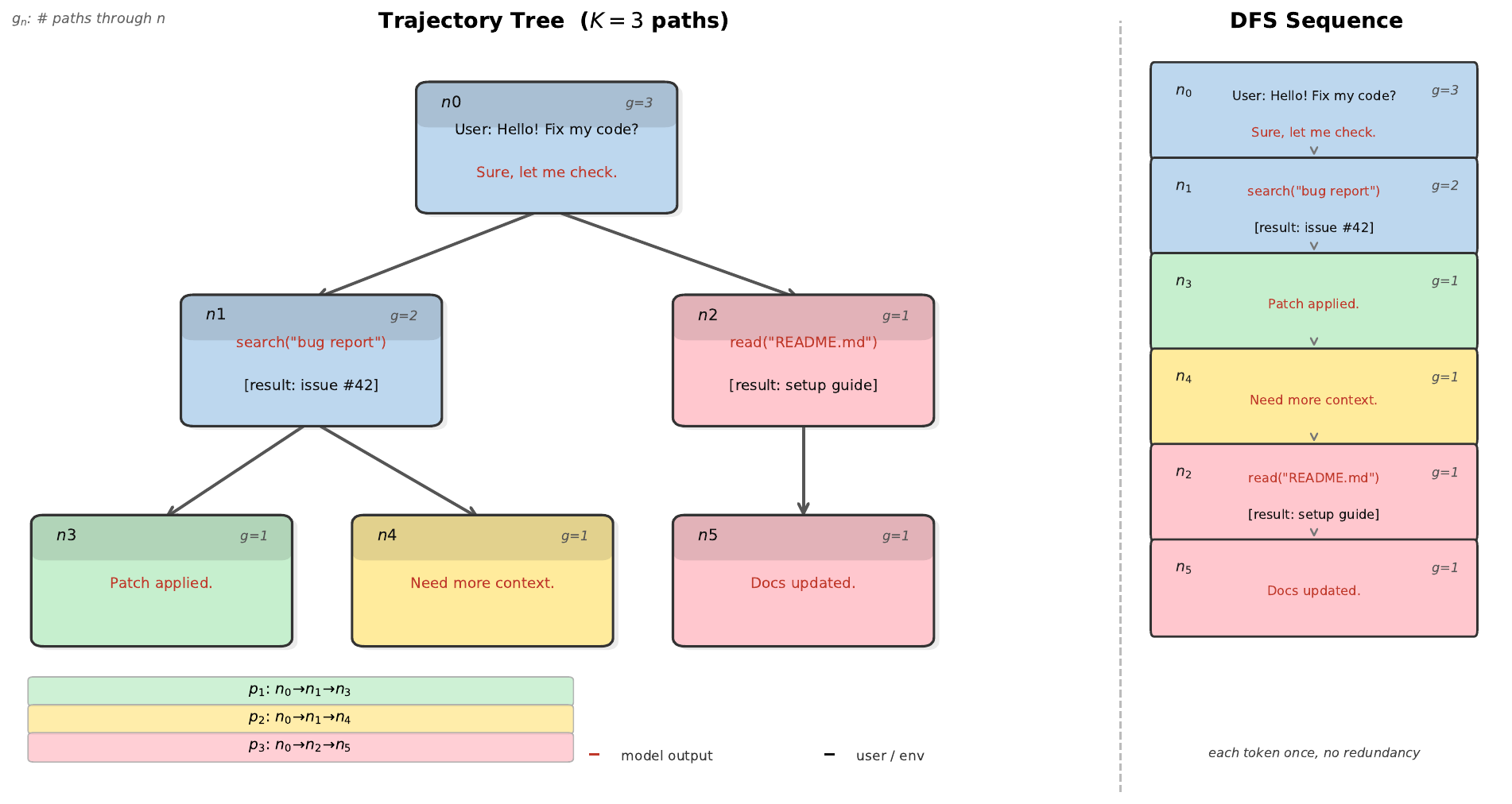}
  \caption{\textbf{Trajectory tree and DFS sequence.}
    Left: a trajectory tree with $K=3$ root-to-leaf paths. Each node $n$ holds a token segment $T_n$; $g_n$ is the number of paths through $n$ (blue nodes are shared prefixes with $g_n>1$). Node text in {\color{red}red} is model output (trained); black is user/environment input. Right: the DFS sequence places all nodes in depth-first order so that every token is computed exactly once, with no redundancy.}
  \label{fig:tree_example}
\end{figure*}

\section{Introduction}
Training Large Language Models (LLMs) in agentic scenarios involves multi-turn interactions with environments, where a single task frequently branches into multiple parallel paths. This branching is driven by tree-structured workflow designs (e.g., tree-structured planning \cite{yao2023tree, zhang2024rest} and sub-agents\cite{kimiteam2026kimik25visualagentic}), concurrent tool invocations \cite{kim2024llmcompilerparallelfunction, abdelaziz-etal-2024-granite}, context management \cite{xu2025mem, packer2024memgptllmsoperatingsystems, zhong2023memorybankenhancinglargelanguage}, as well as runtime behaviors such as think-mode (discarding/replacing intermediate reasoning tokens between turns, e.g. Figure~\ref{fig:tree_example}), and retokenization drift \cite{luo2025agentlightningtrainai}. These mechanisms imply that the context fed into turn $t{+}1$ is often \emph{not} exactly the concatenation of the tokens produced in previous turns (from turn $1$ to turn $t$); therefore, after finishing a task, the collected tokens cannot always be organized as a single list/tensor, but instead form a tree-structured token trajectory with shared prefixes across branches. As a result, training must process multiple sequences with substantial shared prefixes, leading to significant redundant forward/backward computation. Prefix caching~\cite{kwon2023efficientmemorymanagementlarge,pope2022efficientlyscalingtransformerinference} addresses this at inference time by reusing identical KV states for the forward pass, but does not extend to training, where the gradient of each prefix token depends on all branches passing through it.

To address the above challenges, we mathematically derive that the baseline loss---averaging over all branches independently---is algebraically identical to a per-token weighted loss where each token's weight equals the fraction of branches passing through it. The problem therefore reduces to computing the log-probability of every token in the prefix tree exactly once, with no repeated computation across shared prefixes. We propose DFS serialization of the tree, which visits every token exactly once, and adapt full-attention and SSM layers to ensure the resulting log-probabilities match the baseline exactly. An alternative approach is to process the tree one node at a time with differentiable context passing between nodes---this also achieves zero redundant computation but suffers from poor GPU utilisation due to many small forward calls and high kernel-launch overhead (analyzed in Section~\ref{tpack}).

In practice, a single trajectory tree is often too large to fit entirely within GPU memory. We therefore propose \textbf{Redundancy-Free Tree Partitioning}: a hybrid that applies DFS serialization within memory-feasible subtrees for compute efficiency, while using differentiable partition boundaries to relay accumulated states (KV caches and, for SSM hybrid models, recurrent states) without ancestor recomputation, keeping peak memory bounded by a single root-to-leaf path.

Together, DFS serialization with model-layer adaptations and Redundancy-Free Tree Partitioning form \textbf{Tree Training}: an efficient framework for training LLMs on tree-structured trajectories for both SFT and RL, achieving up to \textbf{6.2×} end-to-end training speedup in our experiments.

In summary, our contributions are threefold:

(1) We identify a pervasive yet overlooked property of agentic LLM training: trajectories naturally form overlapping tree-structured prefixes, creating substantial opportunities for computation reuse.

(2) We introduce \textbf{Tree Training}, a paradigm that executes each shared prefix once while achieving mathematically equivalent training to the baseline, via DFS serialization, targeted model-layer adaptations, and---under memory constraints---\textbf{Redundancy-Free Tree Partitioning} with zero redundant computation and peak memory bounded by a single root-to-leaf path.

(3) We demonstrate consistent gains across dense and MoE~\cite{deepseekai2024deepseekv2strongeconomicalefficient} models on diverse datasets, achieving average 6.2× end-to-end speedup---and in agentic RL scenarios---without compromising training fidelity or final model quality.

\section{Related Work}

Prefix caching is widely used to speed up autoregressive decoding by reusing identical prefixes in the forward pass \cite{kwon2023efficientmemorymanagementlarge, pope2022efficientlyscalingtransformerinference, liu2024cachegen, cheng2024large, yao2025cacheblend, cheng2025lmcache, qin2025mooncakekvcachecentricdisaggregatedarchitecture}. For training objectives where the shared part is purely a prompt, batching strategies such as prefix grouping can also reuse the prompt prefix computation without affecting gradients \cite{liu2025prefixgrouperefficientgrpo, wang2024acceleratingdirectpreferenceoptimization}. In addition, Goru et al. \cite{goru2025onepassreasontokenduplication} propose a custom-mask approach to deduplicate repeated reasoning tokens within a single linear conversation. However, it does not address prefix reuse across multiple trajectories, and its prefixes never aggregate loss from multiple branches. As a result, it is a special case of the agentic scenario where the backward pass remains standard.

Prior works have also recognized that agentic/RL generation often forms tree-structured rollouts, and exploit shared prefixes primarily to improve \emph{rollout} efficiency, e.g., tree-search style rollouts and prefix reuse \cite{hou2025treerlllmreinforcementlearning, li2025treepobridginggappolicy, ji2025treesearchllmagent}. However, the subsequent training stage is commonly executed on linearized samples: current training pipelines \cite{luo2025agentlightningtrainai, zhang2025agentrlscalingagenticreinforcement, zhou2025sweetrltrainingmultiturnllm} typically decompose a trajectory tree into separate linear segments and treat each segment independently, causing shared prefixes to be repeatedly recomputed in both forward and backward passes.

Sequence packing \cite{Krell2021SequencePacking} refers to concatenating multiple short sequences into a fixed-length training example to improve training efficiency. Our proposed Tree Partitioning method generalizes this idea: it not only concatenates sequences, but also introduces a strategy for efficiently segmenting overly long training data structured as a prefix tree (noting that a sequence is a special case of a prefix tree). Crucially, by combining Redundancy-Free Tree Partitioning, our method eliminates the boundary recomputation that would otherwise arise from re-including ancestor tokens in each child subtree, thereby ensuring each token is computed exactly once across the entire tree.

Recent hybrid LLM architectures interleave full-attention layers with State Space Model (SSM) layers. Gated Delta Net (GDN)~\cite{yang2025gateddeltanetworksimproving} is a recent SSM variant that improves over earlier SSM architectures such as Mamba~\cite{gu2024mambalineartimesequencemodeling} and Mamba2~\cite{dao2024transformersssmsgeneralizedmodels} by incorporating the delta rule, achieving better associative recall and in-context learning. GDN is adopted as the SSM component in hybrid models such as Qwen3.5~\cite{qwen35blog}. Unlike full-attention layers, SSM layers maintain a recurrent hidden state that is passed sequentially between tokens, which requires dedicated handling when processing tree-structured data.

Overall, unlike our work, these methods primarily target inference-time caching, prompt-only reuse, or single-conversation masking, and do not address the general case where a shared prefix must aggregate gradient contributions from multiple continuations.

\begin{figure*}[t]
  \centering
  \includegraphics[width=0.98\linewidth]{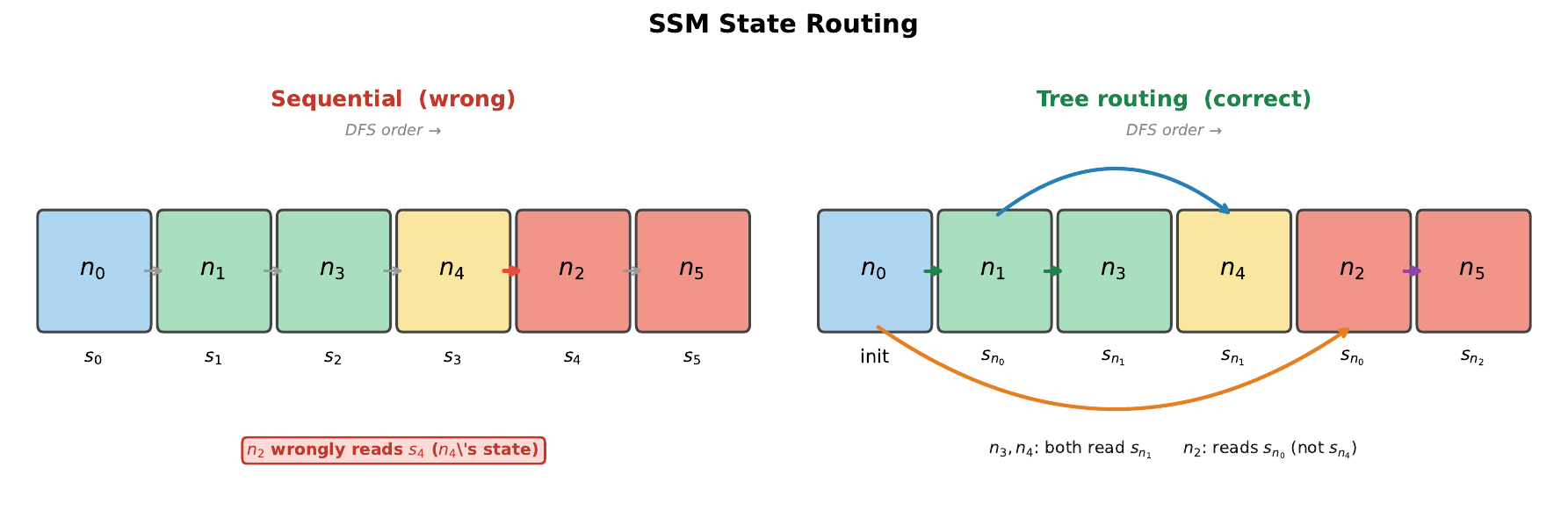}
  \caption{\textbf{SSM state routing.} Sequential flow (left) feeds $n_4$'s state into $n_2$---wrong, since $n_2$ is a sibling of $n_1$, not a descendant of $n_4$. Tree routing (right) routes each chunk to its parent's state: both $n_3$ and $n_4$ read $s_{n_1}$, while $n_2$ correctly reads $s_{n_0}$.}
  \label{fig:ssm_routing}
\end{figure*}

\section{Tree Training} \label{gc}

\subsection{Derivation}\label{dr}

A \textbf{trajectory tree} is a rooted tree in which each node $n$ holds a token segment $T_n$. Each root-to-leaf path spells out a complete trajectory by concatenating the token segments of its nodes in order. Let $\mathcal{T}$ denote the set of all nodes and $\mathcal{P} = \{p_1,\dots,p_K\}$ the set of $K$ root-to-leaf paths (one per leaf). An example with $K=3$ is shown in Figure~\ref{fig:tree_example}.
\textbf{Goal.} Process all tokens in the tree with each token computed \emph{exactly once} (no redundant computation across shared prefixes) in a single forward and backward pass, and produce gradients that are mathematically equivalent to the baseline of running all $K$ paths independently and averaging their losses.

The baseline sep-avg loss runs each of the $K$ paths independently:
\begin{equation}
\label{base_eq}
L_{\text{sep\_avg}} = \frac{1}{K} \sum_{k=1}^{K} \sum_{t \in p_k} \ell_t(\theta)
\end{equation}
where $\ell_t(\theta)$ is any per-token objective.
For SFT, $\ell_t(\theta) = -\log p_\theta(y_t \mid x_{\le t})$;
for RL with policy gradient, $\ell_t(\theta) = -A_t \cdot \log p_\theta(y_t \mid x_{\le t})$
where $A_t$ is the per-token advantage.

Eq.~\eqref{base_eq} iterates over paths then over tokens in each path. We swap the order: iterate over every unique token $t$ and count how many paths include it. Token $t$ belongs to node $n$; let $g_n = |\{p \in \mathcal{P} : n \in p\}|$ be the number of paths passing through $n$, and $g_t \triangleq g_n$ for $t \in T_n$. Token $t$ is therefore counted $g_n$ times in the double sum:
\begin{equation}
\label{swap_eq}
\sum_{k=1}^{K} \sum_{t \in p_k} \ell_t(\theta) \;=\; \sum_{n \in \mathcal{T}} \sum_{t \in T_n} g_n \cdot \ell_t(\theta) \;=\; \sum_{t} g_t \cdot \ell_t(\theta)
\end{equation}
Substituting~\eqref{swap_eq} into~\eqref{base_eq}:
\begin{equation}
\label{sep_pertoken}
L_{\text{sep\_avg}} = \sum_{t} \frac{g_t}{K} \cdot \ell_t(\theta)
\end{equation}
We therefore define the tree loss:
\begin{equation}
\label{tree_loss}
\boxed{L_{\text{tree}} = \sum_{t} \frac{g_t}{K} \cdot \ell_t(\theta)}
\end{equation}
By linearity of differentiation:
\begin{equation}
\frac{\partial L_{\text{tree}}}{\partial \theta} = \sum_{t} \frac{g_t}{K} \cdot \frac{\partial \ell_t}{\partial \theta} {=} \frac{\partial L_{\text{sep\_avg}}}{\partial \theta} \qquad 
\end{equation}
This derivation is model-agnostic and objective-agnostic: since it relies only on the additive structure of $\ell_t$ across tokens and paths---not on its specific form---it applies to any per-token objective. The coefficient $g_t/K$ is merely one special case arising from equal path weights ($w_k = 1/K$), chosen here to simplify the presentation; in general, any path weights $\{w_k\}$ yield a per-token weight $\lambda_t = \sum_{k:\,t \in p_k} w_k$, and the same conclusion holds: the loss reduces to a single weighted sum over unique tokens $\sum_t \lambda_t \cdot \ell_t(\theta)$, achievable in one forward pass by multiplying each token's loss term by $\lambda_t$---without any change to how $\ell_t$ is computed. For example, setting $\lambda_t = 1$ for every unique token gives a valid training objective where each token contributes equally regardless of branch count; it is a distinct objective from sep-avg but equally straightforward to implement. Note that $L_{\text{tree}}$ and $L_{\text{sep\_avg}}$ are identical expressions---the substitution is purely algebraic; it does not yet impose any constraint on how $\ell_t(\theta)$ is computed in practice.

\textbf{Problem reduction.} The goal therefore reduces to: find a method that (1) computes each token \emph{exactly once} with no redundant computation, and (2) ensures that $\ell_t(\theta)$ computed this way matches what it would be in a standalone per-branch forward pass---i.e., \textbf{forward equivalence}:
\begin{equation}
\label{fwd_eq}
\ell_t^{\text{tree}}(\theta) = \ell_t^{p}(\theta) \quad \forall\, p \in \mathrm{paths}(n_t)
\end{equation}
We achieve this via DFS traversal of the token prefix tree, which serializes all nodes in depth-first order so each token appears exactly once, replacing the baseline serialization that repeats prefixes:
\begin{multline}
\label{base_serial}
X_{\text{base}}(i) = \operatorname{Concat}\bigl[\operatorname{token}(i),\, X_{\text{base}}(\operatorname{child}(0,i)),\\
\operatorname{token}(i),\, X_{\text{base}}(\operatorname{child}(1,i)),\; \dots\bigr]
\end{multline}
with:
\begin{multline}
\label{ours_serial}
X_{\text{DFS}}(i) = \operatorname{Concat}\bigl[\operatorname{token}(i),\, X_{\text{DFS}}(\operatorname{child}(0,i)),\\
X_{\text{DFS}}(\operatorname{child}(1,i)),\; \dots\bigr]
\end{multline}
How to enforce~\eqref{fwd_eq} on the \textbf{DFS-serialized} sequence---i.e., ensuring each token's log-probability matches its per-branch value despite the reordering introduced by DFS traversal---is described as follows.

\subsection{Implementation} \label{impl}

\begin{figure}[htb]
  \centering
  \includegraphics[width=0.98\linewidth]{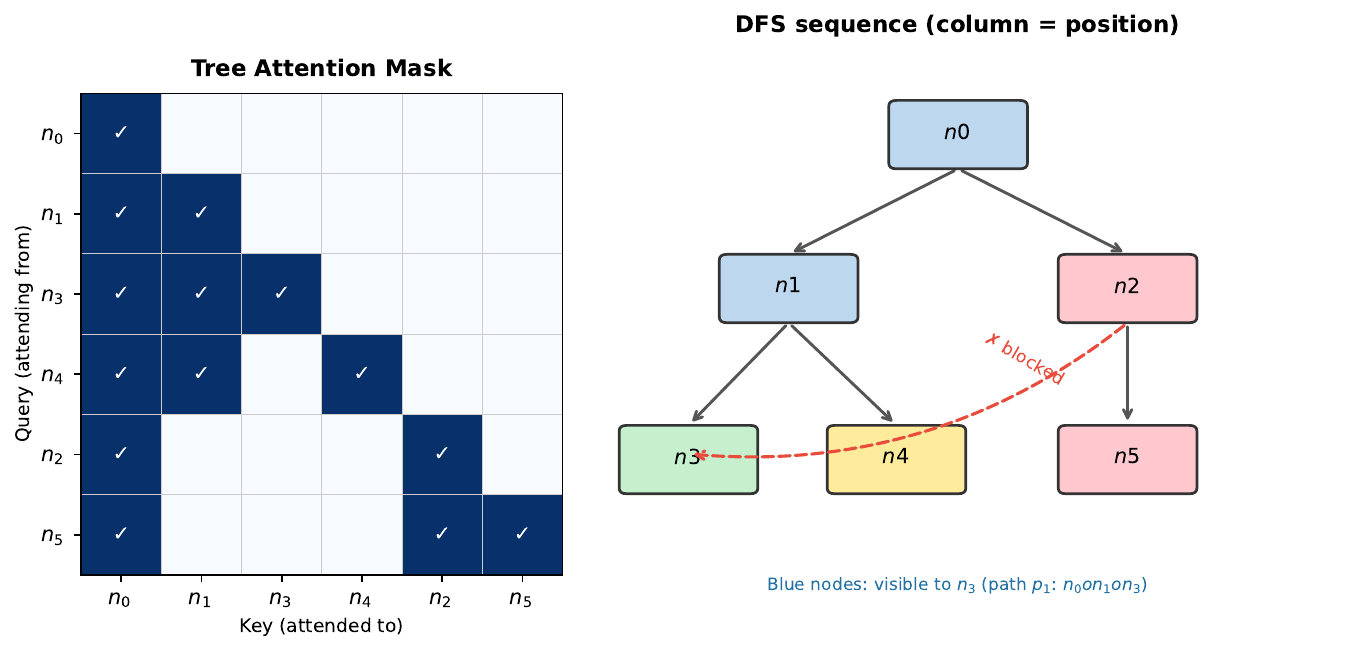}
  \caption{\textbf{Tree attention mask.} Left: the $6\times6$ attention matrix for the DFS sequence; a cell is filled iff the query token can attend to the key token. Tokens from different branches (e.g.\ $n_3$ and $n_2$) are mutually blocked. Right: the corresponding tree structure; the dashed red arrow shows a blocked cross-branch path.}
  \label{fig:attn_mask}
\end{figure}

\paragraph{Attention Mask.}
We construct a tree attention mask such that token $i$ attends to token $j$ iff (1) $j \le i$ and (2) the nodes of $i$ and $j$ lie on a common root-to-leaf path. This prevents cross-branch attention. We implement a GPU kernel based on Flash Attention V3~\cite{shah2024flashattention3fastaccurateattention,wang2025flashmaskefficientrichmask} with node-level shared-prefix masking. The attention pattern is illustrated in Figure~\ref{fig:attn_mask}.

\paragraph{Position Embedding.}
Each token must receive the same position ID as it would in its standalone per-path sequence. A token at offset $j$ within node $n$ is assigned:
\begin{equation}
\mathrm{pos}(t) = \sum_{n' \in \mathrm{anc}(n)\setminus\{n\}} |T_{n'}| + j
\end{equation}
where $\mathrm{anc}(n)$ denotes the ancestors of $n$. Sibling nodes at the same depth share the same position range, matching the per-path baseline and ensuring RoPE~\cite{su2023roformerenhancedtransformerrotary} produces identical results.

\paragraph{SSM Layers (GDN).}
For hybrid models that interleave full-attention with Gated Delta Net (GDN)~\cite{yang2025gateddeltanetworksimproving} SSM layers, the attention mask alone is insufficient: GDN layers maintain a recurrent hidden state passed sequentially between chunk boundaries, so a na\"{i}ve DFS layout would carry the state from one branch into the next sibling branch. We apply two fixes. We treat each tree node's token segment as one \emph{chunk}---the unit of SSM state transfer. (i)~\textbf{State routing}: each chunk $c$ reads its initial SSM state from its \emph{parent} chunk $\pi(c)$ rather than the preceding chunk in DFS order:
\begin{equation}
\mathbf{h}_c^{(0)} = \mathbf{h}_{\pi(c)}^{(\mathrm{end})}, \qquad \mathbf{h}_{-1}^{(\mathrm{end})} \triangleq \mathbf{0}
\end{equation}
DFS order guarantees $\pi(c)<c$, so parent states are always ready. Sibling chunks share the same parent state tensor, so their gradients accumulate there automatically during the backward pass. Figure~\ref{fig:ssm_routing} illustrates the difference between sequential and tree state routing.

\begin{figure}[htb]
  \centering
  \includegraphics[width=0.98\linewidth]{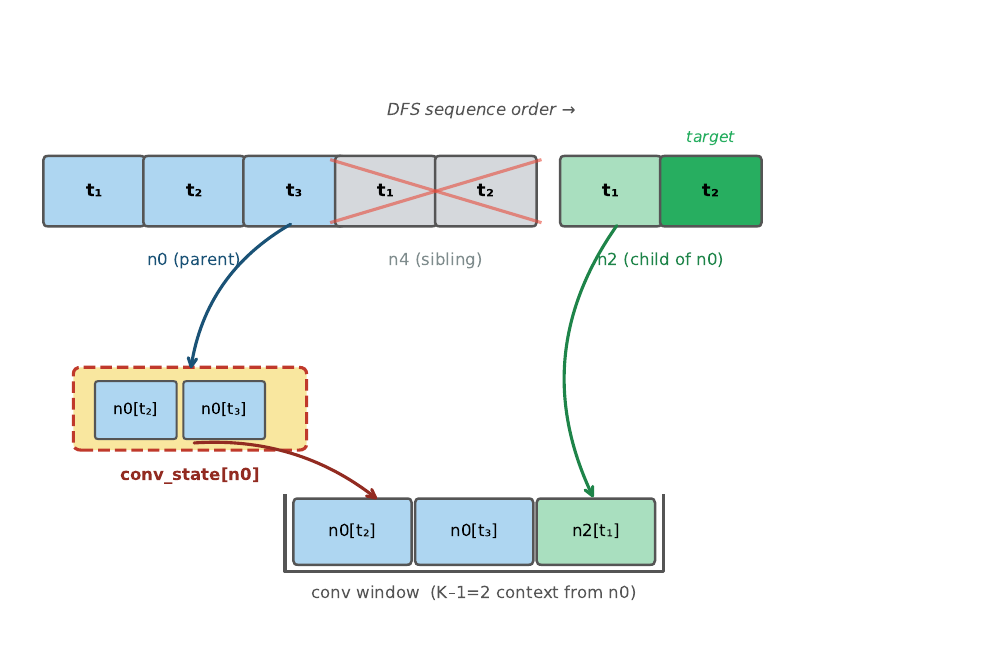}
  \caption{\textbf{SSM causal convolution} ($K_{\mathrm{conv}}=3$). When processing $n_2$ (child of $n_0$), the correct conv window is formed by prepending \texttt{conv\_state}[$n_0$]---the stored last $K_{\mathrm{conv}}{-}1$ effective tokens of the parent---as left context, bypassing the DFS-adjacent sibling tokens ($n_4$) that occupy the preceding positions.}
  \label{fig:ssm_conv}
\end{figure}

(ii)~\textbf{Causal convolution}: SSM layers include a causal conv1d with kernel size $K_{\mathrm{conv}}$, which requires the preceding $K_{\mathrm{conv}}{-}1$ tokens as left context. In the DFS sequence, the tokens immediately before a child chunk belong to a sibling branch rather than the ancestor path, so using them as context is incorrect. Instead, each chunk $c$ uses $\mathbf{ctx}_{\pi(c)}$---the last $K_{\mathrm{conv}}{-}1$ output tokens of its parent chunk $\pi(c)$, saved after processing the parent---as its left context:
\begin{equation}
\mathbf{y}_c = \mathrm{Conv}\!\left([\,\mathbf{ctx}_{\pi(c)}\,;\,\mathbf{x}_c\,]\right)_{[K_{\mathrm{conv}}-1:]}
\end{equation}
This replicates the context each token would see in an independent per-branch forward pass. Figure~\ref{fig:ssm_conv} illustrates this for $K_{\mathrm{conv}}=3$.

\paragraph{Loss Weights.}
For each token $t$ in the DFS-serialized sequence, its per-token loss $\ell_t(\theta)$ is
multiplied by a weight $\lambda_t$ before reduction.
The specific value of $\lambda_t$ depends on the training objective: for the sep-avg
baseline with equal path weights this gives $\lambda_t = g_t/K$ as derived in
Eq.~\eqref{tree_loss}; in general $\lambda_t$ can be any value determined by the
path weighting scheme (see Section~\ref{dr}).
This is applied as a single element-wise multiplication on the per-token loss
tensor---requiring no modification to the backward pass and negligible overhead.
Full implementation details and float32 gradient verification appear in
Appendix~\ref{app:dfs_impl}.

\subsection{Redundancy-Free Tree Partitioning} \label{tpack}

\begin{figure}[htb]
  \centering
  \includegraphics[width=0.98\linewidth]{img/dfspack.pdf}
  \caption{Memory-constrained training on a tree with 83k unique tokens (GPU limit $C=60\text{k}$). Baseline flattening: 164k tokens. Standard Tree Partitioning (no differentiable boundaries): 102k tokens due to boundary recomputation. With differentiable partition boundaries, boundary recomputation is eliminated entirely: 83k tokens, equal to the tree's unique token count.}
  \label{tree_pack_img}
\end{figure}

When $N_{\text{tree}}$ exceeds GPU memory capacity, the tree must be split into subtrees,
each processed as a single DFS forward pass.
Each subtree should be as large as GPU memory allows: larger subtrees pack more tokens
into each forward call, improving compute density and reducing the number of model
invocations (see the efficiency analysis in the last paragraph of this section).

Two design questions arise:
(1)~how to partition the tree optimally while keeping peak activation memory bounded; and
(2)~how to avoid redundant computation at partition boundaries---adjacent subtrees share
a common prefix, and without a dedicated mechanism each child partition re-includes those
shared ancestor tokens in its own DFS sequence, introducing redundant computation
(e.g. Figure~\ref{tree_pack_img}).

\paragraph{Partitioning.}
Partitions must follow tree node boundaries: each partition is a connected subtree, and
every cut occurs at a node boundary.
The reason is peak activation memory.
The backward pass chains gradients bottom-up through the partition dependency graph;
to bound peak memory at $\mathcal{O}(\text{max-path tokens})$, this graph must itself be
a tree---meaning each partition has exactly one parent partition.
This holds if and only if cuts are at node boundaries.
Arbitrary cuts would place tokens from independent subtrees into the same partition;
its backward must then simultaneously retain the computation graphs of \emph{multiple}
parent partitions (and transitively their ancestors), potentially keeping the graphs of
all ancestor subtrees of all branches in memory at once.

The optimization goal is therefore to minimise $N_{\text{partitions}}$---equivalently,
maximise token utilisation of each partition---subject to each partition containing at
most $C$ tokens.
This is a standard bin packing problem on tree subgraphs, solvable with general-purpose
combinatorial solvers such as OR-Tools~\cite{ortools}.

\paragraph{Differentiable partition boundaries.}
The second problem---avoiding redundant recomputation of shared prefixes across
subtrees---is solved by making partition boundaries differentiable.
At each cut node $n_c$, the parent partition's accumulated KV tensors (and SSM states
for hybrid models) are detached and re-exposed as differentiable leaf variables;
the child partition attends to these cached ancestor
key-value pairs without re-executing any ancestor token.
After the child partition's backward, gradients in the leaf tensors are passed back into
the parent partition's retained computation graph via an explicit backward pass through
the parent---the same gradient-relay mechanism used in pipeline-parallel training---so each token is
computed exactly once and total computation equals $N_{\text{tree}}$
(Figure~\ref{tree_pack_img}: 102k $\to$ 83k).
For SSM hybrid models, the gateway additionally carries the recurrent hidden state and
causal-conv context at the cut node, chained back identically.
Details (ancestor-aware attention mask, depth-based position offset,
float32 accumulation for multi-child branching, SSM state injection) are listed in Appendix~\ref{app:hybrid_impl}.

\paragraph{DFS packing versus per-node processing.}
The differentiable boundary mechanism above can be applied at \emph{every} tree node
boundary---not just partition boundaries---yielding \emph{per-node processing}: the model
is called $N_{\text{nodes}}$ times, each processing one node's own token segment, with the
accumulated KV relayed from the parent node.
Both approaches achieve zero redundant computation; DFS packing is preferable for two reasons.
First, DFS processes all $N_{\text{tree}}$ tokens in $\mathcal{O}(N_{\text{partitions}})$
calls instead of $N_{\text{nodes}}$, directly reducing CPU/kernel-launch overhead.
Second, each DFS call covers far more tokens, making operations such as linear projections
(GEMMs) significantly denser: in per-node processing, each call reloads the weight matrix
with only the current node's tokens to amortise it, while DFS amortises the same weight
load over the entire partition.
For attention, the same principle holds: in per-node processing, attention intensity is
bounded by the node token count with no fixed cost to amortise, while DFS processes the
full tree depth in one call.
For SSM hybrid models, in per-node processing the recurrent state is written to and reloaded
from HBM at every node boundary; DFS keeps it on-chip throughout the partition, avoiding
the dominant memory bottleneck of SSM layers entirely~\cite{gupta2026persistent}.

\begin{figure*}[t]
  \centering
  \hspace{0.035\linewidth}
  \includegraphics[width=0.85\linewidth]{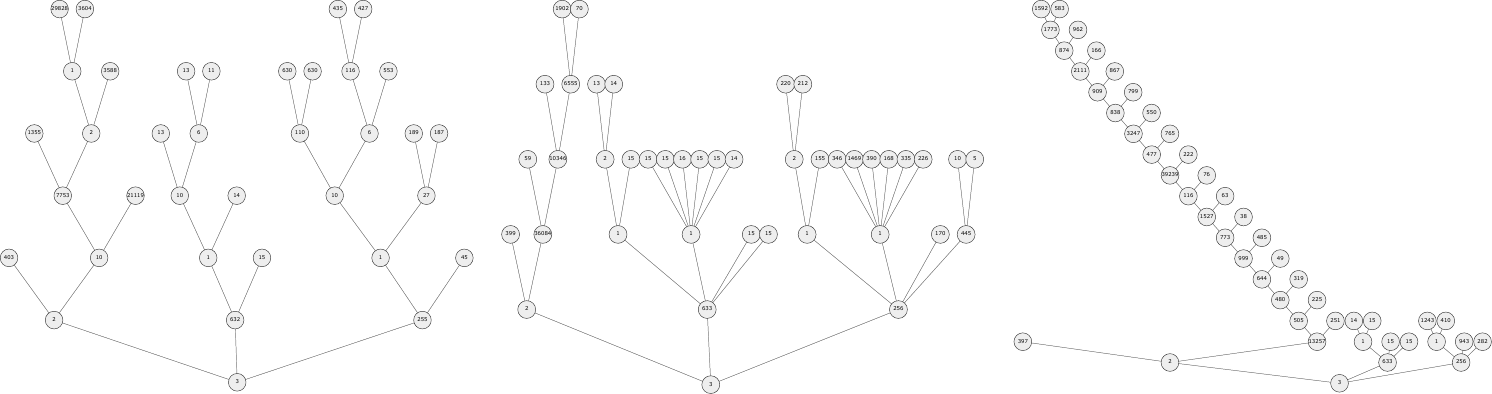}
  \vspace{0.5em}
  \includegraphics[width=0.95\linewidth]{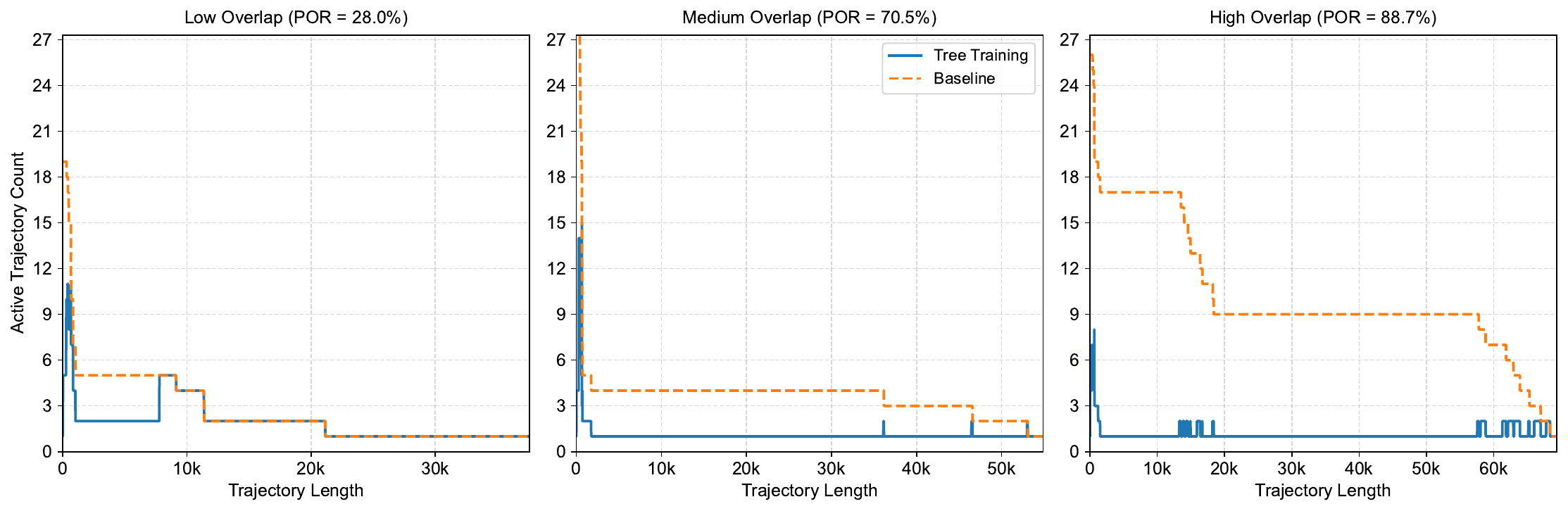}
  \caption{
    \textbf{Real agentic trajectory trees and their overlap characteristics.}
    The upper row shows representative trees from multi-turn agentic RL rollouts with \textbf{Low}, \textbf{Medium}, and \textbf{High Overlap}.
    The lower row plots \textit{active trajectory counts} over length; the area ratio reflects the theoretical token reuse ratio.
    }
  \label{fig:data_and_por}
  \vspace{-0.5em}
\end{figure*}

\begin{figure*}[ht]
  \centering
    \includegraphics[width=0.48\textwidth]{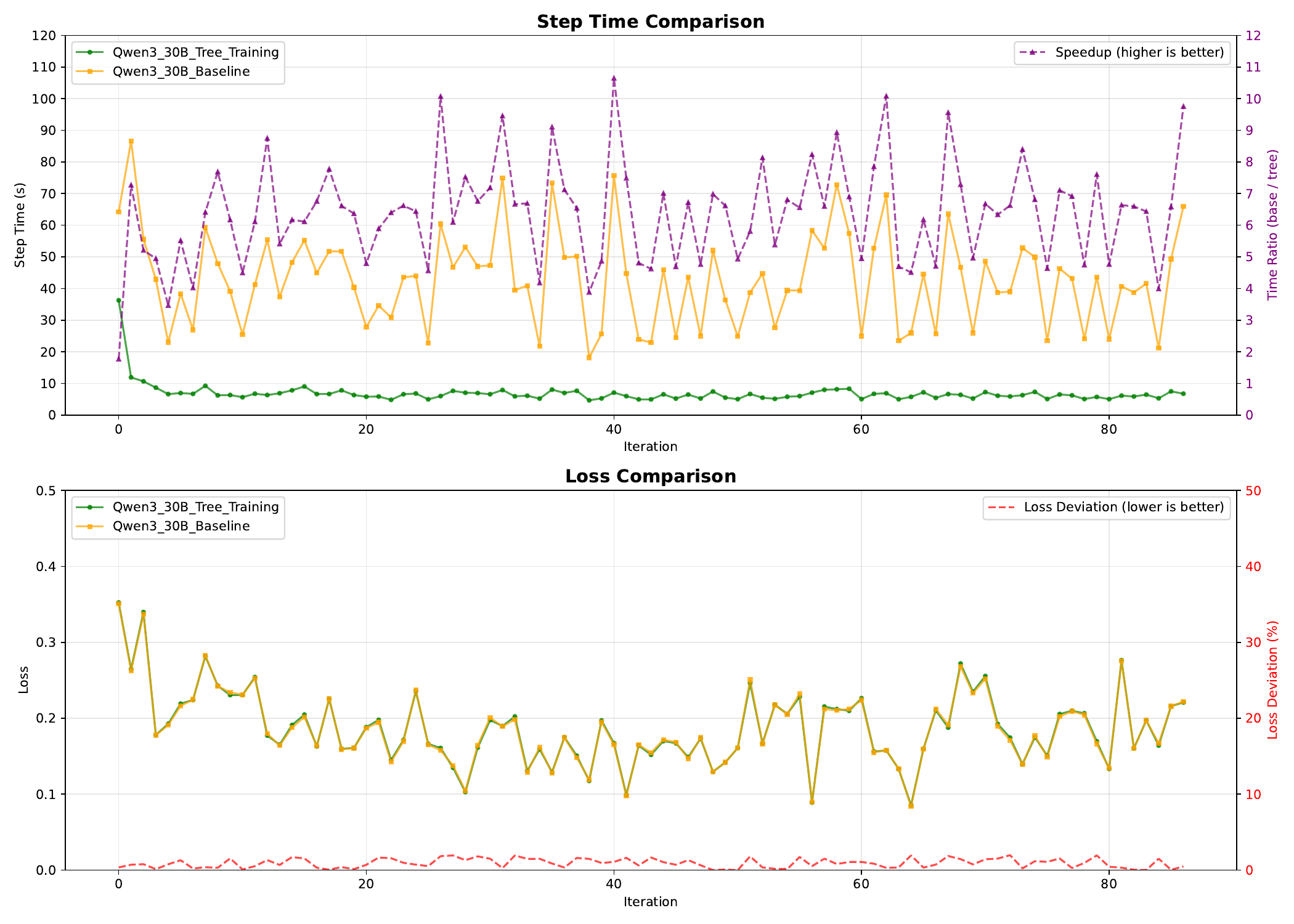}
    \hfill
    \includegraphics[width=0.48\textwidth]{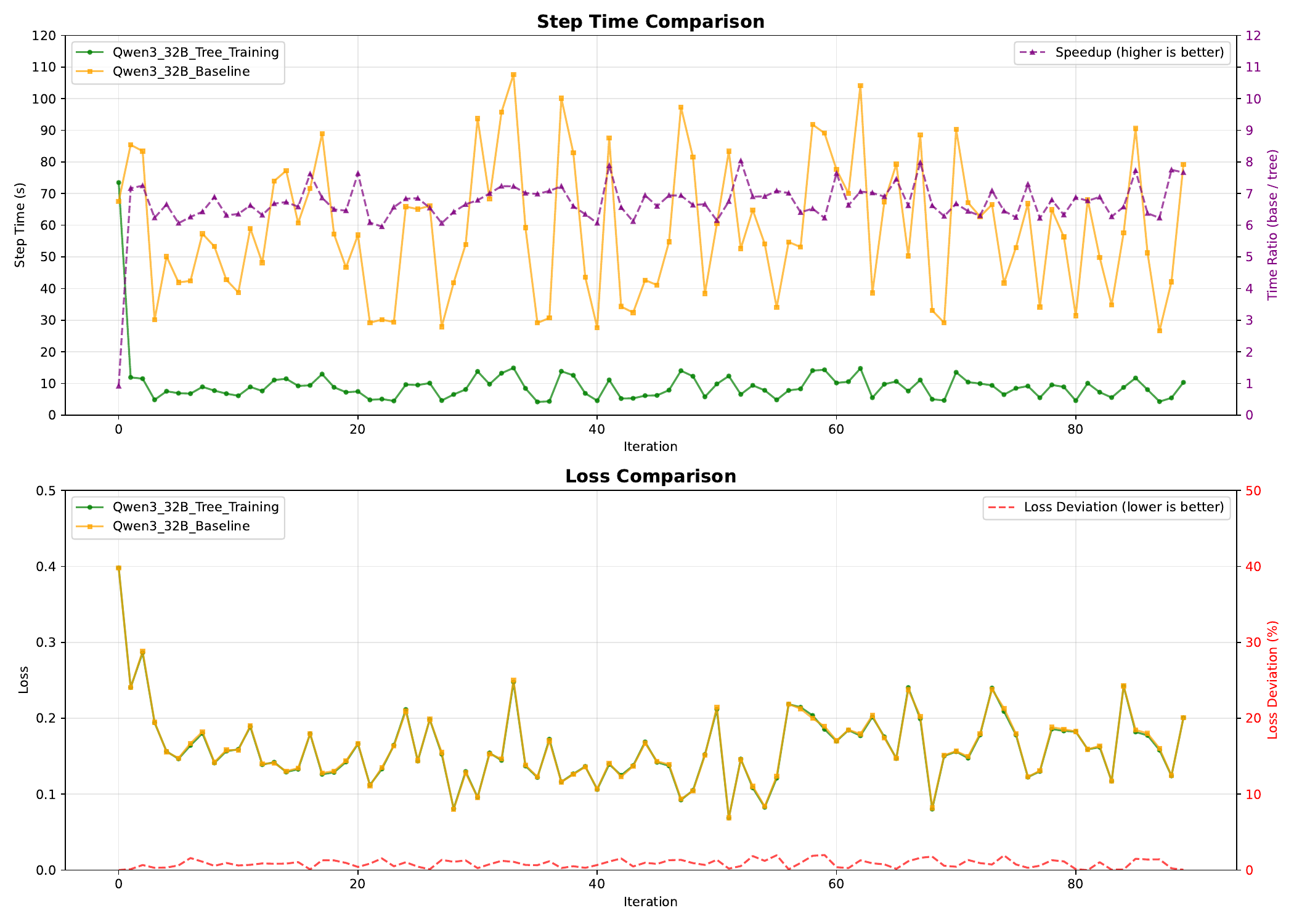}
  \caption{
  \textbf{
  End-to-end training speedup and loss comparison using real world rollout data with think mode turned on.} 
  Results for the MoE model Qwen3-30B~\cite{yang2025qwen3technicalreport} are shown on the left, and for the dense model Qwen3-32B on the right. For each model, the top plot illustrates the training speedup achieved by Tree Training, while the bottom plot displays the average relative error in loss.
  }
  \vspace{-1em}
  \label{fig:real_agentic_data_end_to_end_speedup}
\end{figure*}

\subsection{Does Tree Training introduce bias in the gradient updates?}

Tree Training does not introduce bias during training. Each global batch is a self-contained tree generated from a single rollout---the prefix-tree structure arises naturally within one task, so tree partitioning is organized and processed entirely within one gradient accumulation step, never mixing trajectories from different global batches. The data shuffle operation is applied only between complete tree samples and never disrupts the internal tree structure.

\section{Experiments}
We evaluate Tree Training along three dimensions: gradient correctness, efficiency under realistic memory constraints, and end-to-end training performance. Across all settings, reusing shared prefixes from overlapping trajectories substantially reduces redundant computation while preserving equivalence to standard sequence-based baselines.

\subsection{Metrics}
\label{exp_setting}

%\paragraph{Environment}
To clearly separate dataset structure from runtime effects, we define the following metric.

%The \textbf{Potential Overlap Ratio (POR)} is defined as the ratio between the number of prefix tokens and the total number of tokens in each trajectory prefix tree, averaged over the entire dataset, which represents the \textit{theoretical upper bound} of computation reuse if memory were unlimited.

\textbf{Potential Overlap Ratio (POR)} quantifies the theoretical upper bound of computation reuse by measuring the redundancy inherent in shared prefixes. We define POR as the ratio of overlapping tokens to the total token count of the flattened dataset:
\begin{equation}
\text{POR} = 1 - \frac{N_{\text{tree}}}{N_{X_{base}(root)}}
\end{equation}
Here, $N_{\text{tree}}$ denotes the token count in the prefix tree, while $N_{X_{base}(root)}$ represents the token counts of the serialized tree produced by the baseline method described in Equation~\eqref{base_eq}. Using the example from Figure~\ref{tree_pack_img}, the POR is calculated as $1 - \frac{83\text{k}}{164\text{k}} \approx \mathbf{49.4\%}$, indicating that nearly half of the baseline computation is redundant.
Since Redundancy-Free Tree Partitioning computes each token exactly once regardless of memory constraints, the theoretical speedup upper bound is $1/(1-\text{POR})$.

\subsection{Evaluation Setups}

\begin{figure}[t]
  \centering
  \includegraphics[width=0.85\linewidth]
{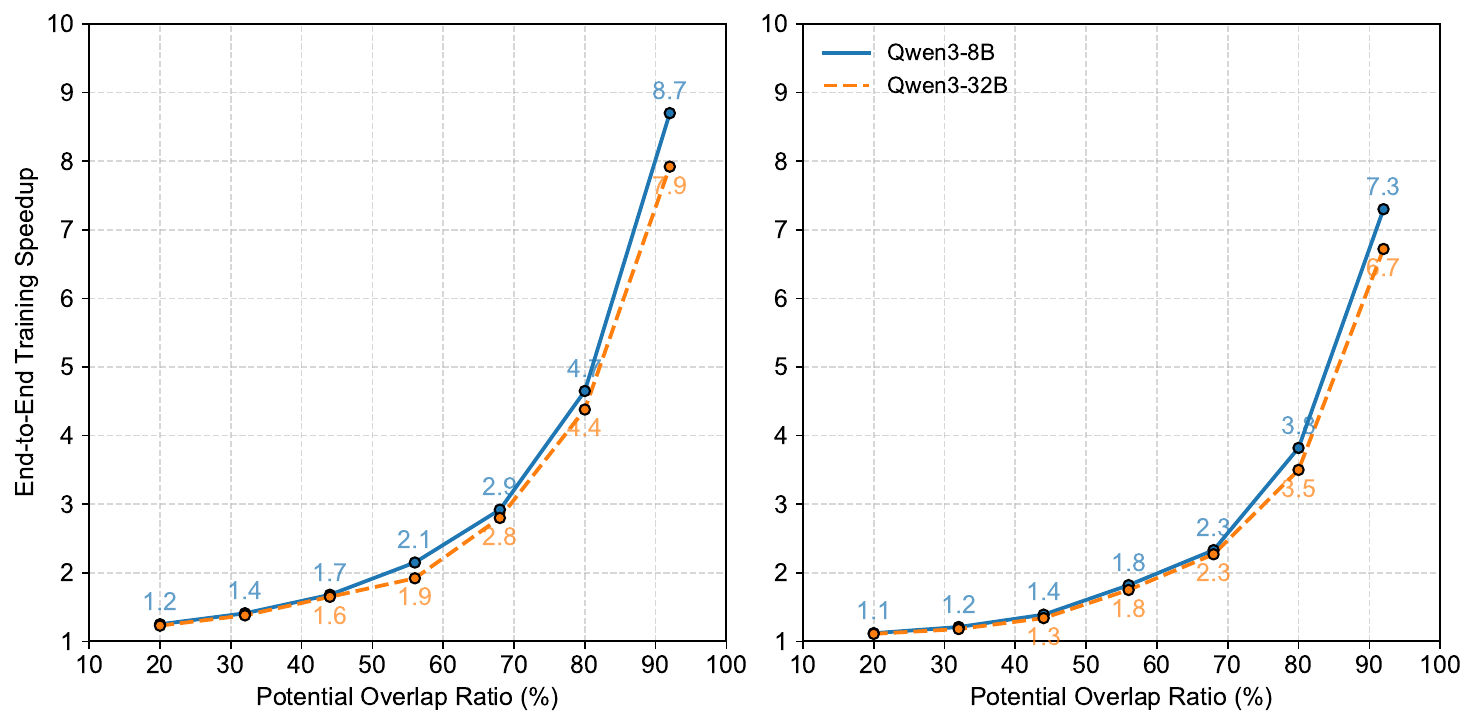}
  \caption{
  \textbf{
  End-to-end training speedup of Tree Training across datasets with varying POR.} 
  Each subfigure reports the relative reduction in total training time of Tree Training compared to the baseline. (a) synthetic datasets where the full tree fits in GPU memory, (b) synthetic datasets requiring tree partitioning under memory constraints.
  }
  \vspace{-1em}
  \label{fig:end_to_end_speedup}
\end{figure}

We evaluate the end-to-end training efficiency by comparing the total time required to process equivalent data between Tree Training and the baseline described in Section~\ref{gc}. This baseline aligns with the current standard practice for training on branched trajectories, as most existing training frameworks only support taking batch of \textbf{list} (tensor) as training inputs.

% Experiments are performed on three dataset settings: synthetic datasets where the full trajectory tree fits in GPU memory, synthetic datasets requiring tree partitioning under memory limits, and real agentic RL rollouts collected from multi-turn reasoning and dialogue tasks. 

All experiments were conducted on clusters of 64 NVIDIA Hopper GPUs using Megatron-Core \cite{shoeybi2020megatronlmtrainingmultibillionparameter} for distributed training. 
Sequence Packing \cite{Krell2021SequencePacking} is used in our baseline setting, representing the standard batching strategy used in large-scale LLM fine-tuning and RL training.

\subsection{Numerical Precision}
While the derivation establishes exact mathematical equivalence, hardware floating-point arithmetic introduces unavoidable small numerical deviations in practice. The DFS-serialized forward pass accumulates intermediate results in a different order than independent per-branch passes, and cuBLAS may select different GEMM kernels for different tensor shapes, both causing non-associativity errors inherent to finite-precision floating-point arithmetic. These are not implementation bugs but fundamental hardware-level effects, and as shown in the next section, they do not affect training dynamics in practice.

\subsection{Speedup and Correctness on real-world agentic trajectories}
To assess real-world applicability, we collect agentic RL rollouts from multi-turn reinforcement learning tasks. Figure~\ref{fig:data_and_por} presents representative realistic agentic trajectory trees (top) alongside their corresponding token-overlap distributions (bottom). These trajectories are collected from software engineering tasks (SWE-smith~\cite{yang2025swesmithscalingdatasoftware}), executed via agentic scaffolds Claude Code~\cite{cc}. Most branching behaviors of the left two trees are caused by concurrent tool execution and retokenization drift \cite{luo2025agentlightningtrainai}, while the right tree is caused by the think mode.
These examples illustrate a broad spectrum of prefix sharing, ranging from low to high POR (from 28.0\% to 88.7\%), and reveal that real agentic datasets tend to exhibit sparse, unbalanced trees with highly variable reuse across different depths and branching factors. 

To validate the correctness and speedup of our method, we illustrate the loss comparison and step-time comparison on real-world agentic trajectories as shown in Figure~\ref{fig:real_agentic_data_end_to_end_speedup}:

(1) Speedup (Top Panel): The theoretical speedup upper bound $\frac{1}{(1-\text{POR})}$ for this dataset is \textbf{6.5×}. We achieved average realized speedups of \textbf{6.3×} (32B Dense) and \textbf{6.2×} (30B MoE). This demonstrates that our system incurs negligible overhead, capturing over \textbf{95\%} of the theoretical potential. The step-wise fluctuation (2×–10×) faithfully reflects the varying intrinsic POR of dynamic tree samples.

(2) Correctness (Bottom Panel): Tree Training reproduces the baseline’s loss and gradient behaviors, with numerical deviations consistently maintained within acceptable precision tolerance. The training loss curves for Baseline (Yellow) and Ours (Green) overlap perfectly. The deviation (Red dashed) is negligible (<1\%), confirming strict mathematical equivalence.

\subsection{Speedup of datasets with varying POR}

For controlled experiments, we construct synthetic datasets with POR values from 20\% to 92\%, while keeping the number of leaf trajectories and total tokens constant across settings. This setup allows us to systematically study how the overlap degree influences computation reuse and throughput. 

As shown in Figure~\ref{fig:end_to_end_speedup}, Tree Training consistently reduces total training time across all conditions, with the improvement magnitude increasing monotonically with the Potential Overlap Ratio (POR). 
In the ideal setting where full trees fit in memory, Tree Training achieves up to 8.7× speedup by eliminating redundant computation through shared prefix reuse. 
%Under realistic memory constraints, tree partitioning introduces moderate overhead but still yields 4.5–5.3× acceleration. 

\subsection{Memory footprint}
In our implementation, even when large trees must be split (achieved via Redundancy-Free Tree Partitioning), our method eliminates all redundant computation and our performance closely tracks the theoretical speedup upper bound of $\frac{1}{(1-\text{POR})}$, demonstrating that the gateway mechanism incurs negligible runtime overhead. The additional tensors required by Tree Training total only \textbf{1.2 MB} on Qwen3-32B~\cite{yang2025qwen3technicalreport}, compared to the 64000 MB minimum activation memory, confirming negligible memory overhead.

\subsection{The gain of training on all tokens of trajectory trees}
Training on all tokens of trajectory trees, compared to training on only a single trajectory (in our baseline experiment, we select the longest trajectory as common practice), leads to a clear performance gain, especially for thinking models. 

On Terminal Bench 2.0\cite{merrill2026terminalbenchbenchmarkingagentshard}, we use a cold-started Qwen3-32B~\cite{yang2025qwen3technicalreport} as the base checkpoint for RL training. The model trained on the full tree (with think-mode turned on) achieves a score (avg@4) of 28.8, whereas the baseline score is 20.9.

% As shown in Figure~\ref{fig:real_agentic_data_end_to_end_speedup}, extensive experiments on real-world rollout data demonstrate that our approach maintains a loss alignment within 1\% relative error while achieving average 6.2× training speedup based on both dense model(qwen3-32B) and moe model(qwen3-30B-A3B), confirming its practical efficacy.

\section{Conclusion}

We propose Tree Training, a performance optimization framework during the training phase. It aims to minimize redundant computation and boost efficiency while strictly maintaining the gradient values of the parameters undergoing optimization. In contrast to existing prefix reuse schemes, our Tree Training is not only suitable when the prefix is just a prompt that does not contribute to gradients, but also when the prefix is involved in gradient updates.

\clearpage

\bibliography{main}

@misc{hou2025treerlllmreinforcementlearning,
      title={TreeRL: LLM Reinforcement Learning with On-Policy Tree Search}, 
      author={Zhenyu Hou and Ziniu Hu and Yujiang Li and Rui Lu and Jie Tang and Yuxiao Dong},
      year={2025},
      eprint={2506.11902},
      archivePrefix={arXiv},
      primaryClass={cs.LG},
      url={https://arxiv.org/abs/2506.11902}, 
}

@misc{li2025treepobridginggappolicy,
      title={TreePO: Bridging the Gap of Policy Optimization and Efficacy and Inference Efficiency with Heuristic Tree-based Modeling}, 
      author={Yizhi Li and Qingshui Gu and Zhoufutu Wen and Ziniu Li and Tianshun Xing and Shuyue Guo and Tianyu Zheng and Xin Zhou and Xingwei Qu and Wangchunshu Zhou and Zheng Zhang and Wei Shen and Qian Liu and Chenghua Lin and Jian Yang and Ge Zhang and Wenhao Huang},
      year={2025},
      eprint={2508.17445},
      archivePrefix={arXiv},
      primaryClass={cs.LG},
      url={https://arxiv.org/abs/2508.17445}, 
}

@misc{ji2025treesearchllmagent,
      title={Tree Search for LLM Agent Reinforcement Learning}, 
      author={Yuxiang Ji and Ziyu Ma and Yong Wang and Guanhua Chen and Xiangxiang Chu and Liaoni Wu},
      year={2025},
      eprint={2509.21240},
      archivePrefix={arXiv},
      primaryClass={cs.LG},
      url={https://arxiv.org/abs/2509.21240}, 
}

@misc{liu2025prefixgrouperefficientgrpo,
      title={Prefix Grouper: Efficient GRPO Training through Shared-Prefix Forward}, 
      author={Zikang Liu and Tongtian Yue and Yepeng Tang and Longteng Guo and Junxian Cai and Qingbin Liu and Xi Chen and Jing Liu},
      year={2025},
      eprint={2506.05433},
      archivePrefix={arXiv},
      primaryClass={cs.LG},
      url={https://arxiv.org/abs/2506.05433}, 
}

@misc{wang2024acceleratingdirectpreferenceoptimization,
      title={Accelerating Direct Preference Optimization with Prefix Sharing}, 
      author={Franklin Wang and Sumanth Hegde},
      year={2024},
      eprint={2410.20305},
      archivePrefix={arXiv},
      primaryClass={cs.LG},
      url={https://arxiv.org/abs/2410.20305}, 
}

@misc{kwon2023efficientmemorymanagementlarge,
      title={Efficient Memory Management for Large Language Model Serving with PagedAttention}, 
      author={Woosuk Kwon and Zhuohan Li and Siyuan Zhuang and Ying Sheng and Lianmin Zheng and Cody Hao Yu and Joseph E. Gonzalez and Hao Zhang and Ion Stoica},
      year={2023},
      eprint={2309.06180},
      archivePrefix={arXiv},
      primaryClass={cs.LG},
      url={https://arxiv.org/abs/2309.06180}, 
}

@inproceedings{liu2024cachegen,
  title={Cachegen: Kv cache compression and streaming for fast large language model serving},
  author={Liu, Yuhan and Li, Hanchen and Cheng, Yihua and Ray, Siddhant and Huang, Yuyang and Zhang, Qizheng and Du, Kuntai and Yao, Jiayi and Lu, Shan and Ananthanarayanan, Ganesh and others},
  booktitle={Proceedings of the ACM SIGCOMM 2024 Conference},
  pages={38--56},
  year={2024}
}

@article{cheng2024large,
  title={Do Large Language Models Need a Content Delivery Network?},
  author={Cheng, Yihua and Du, Kuntai and Yao, Jiayi and Jiang, Junchen},
  journal={arXiv preprint arXiv:2409.13761},
  year={2024}
}

@inproceedings{yao2025cacheblend,
  author = {Yao, Jiayi and Li, Hanchen and Liu, Yuhan and Ray, Siddhant and Cheng, Yihua and Zhang, Qizheng and Du, Kuntai and Lu, Shan and Jiang, Junchen},
  title = {CacheBlend: Fast Large Language Model Serving for RAG with Cached Knowledge Fusion},
  year = {2025},
  doi = {10.1145/3689031.3696098},
  booktitle = {Proceedings of the Twentieth European Conference on Computer Systems},
  pages = {94--109}
}

@misc{kimiteam2026kimik25visualagentic,
      title={Kimi K2.5: Visual Agentic Intelligence}, 
      author={Kimi Team and Tongtong Bai and Yifan Bai and Yiping Bao and S. H. Cai and Yuan Cao and Y. Charles and H. S. Che and Cheng Chen and Guanduo Chen and Huarong Chen and Jia Chen and Jiahao Chen and Jianlong Chen and Jun Chen and Kefan Chen and Liang Chen and Ruijue Chen and Xinhao Chen and Yanru Chen and Yanxu Chen and Yicun Chen and Yimin Chen and Yingjiang Chen and Yuankun Chen and Yujie Chen and Yutian Chen and Zhirong Chen and Ziwei Chen and Dazhi Cheng and Minghan Chu and Jialei Cui and Jiaqi Deng and Muxi Diao and Hao Ding and Mengfan Dong and Mengnan Dong and Yuxin Dong and Yuhao Dong and Angang Du and Chenzhuang Du and Dikang Du and Lingxiao Du and Yulun Du and Yu Fan and Shengjun Fang and Qiulin Feng and Yichen Feng and Garimugai Fu and Kelin Fu and Hongcheng Gao and Tong Gao and Yuyao Ge and Shangyi Geng and Chengyang Gong and Xiaochen Gong and Zhuoma Gongque and Qizheng Gu and Xinran Gu and Yicheng Gu and Longyu Guan and Yuanying Guo and Xiaoru Hao and Weiran He and Wenyang He and Yunjia He and Chao Hong and Hao Hu and Jiaxi Hu and Yangyang Hu and Zhenxing Hu and Ke Huang and Ruiyuan Huang and Weixiao Huang and Zhiqi Huang and Tao Jiang and Zhejun Jiang and Xinyi Jin and Yu Jing and Guokun Lai and Aidi Li and C. Li and Cheng Li and Fang Li and Guanghe Li and Guanyu Li and Haitao Li and Haoyang Li and Jia Li and Jingwei Li and Junxiong Li and Lincan Li and Mo Li and Weihong Li and Wentao Li and Xinhang Li and Xinhao Li and Yang Li and Yanhao Li and Yiwei Li and Yuxiao Li and Zhaowei Li and Zheming Li and Weilong Liao and Jiawei Lin and Xiaohan Lin and Zhishan Lin and Zichao Lin and Cheng Liu and Chenyu Liu and Hongzhang Liu and Liang Liu and Shaowei Liu and Shudong Liu and Shuran Liu and Tianwei Liu and Tianyu Liu and Weizhou Liu and Xiangyan Liu and Yangyang Liu and Yanming Liu and Yibo Liu and Yuanxin Liu and Yue Liu and Zhengying Liu and Zhongnuo Liu and Enzhe Lu and Haoyu Lu and Zhiyuan Lu and Junyu Luo and Tongxu Luo and Yashuo Luo and Long Ma and Yingwei Ma and Shaoguang Mao and Yuan Mei and Xin Men and Fanqing Meng and Zhiyong Meng and Yibo Miao and Minqing Ni and Kun Ouyang and Siyuan Pan and Bo Pang and Yuchao Qian and Ruoyu Qin and Zeyu Qin and Jiezhong Qiu and Bowen Qu and Zeyu Shang and Youbo Shao and Tianxiao Shen and Zhennan Shen and Juanfeng Shi and Lidong Shi and Shengyuan Shi and Feifan Song and Pengwei Song and Tianhui Song and Xiaoxi Song and Hongjin Su and Jianlin Su and Zhaochen Su and Lin Sui and Jinsong Sun and Junyao Sun and Tongyu Sun and Flood Sung and Yunpeng Tai and Chuning Tang and Heyi Tang and Xiaojuan Tang and Zhengyang Tang and Jiawen Tao and Shiyuan Teng and Chaoran Tian and Pengfei Tian and Ao Wang and Bowen Wang and Chensi Wang and Chuang Wang and Congcong Wang and Dingkun Wang and Dinglu Wang and Dongliang Wang and Feng Wang and Hailong Wang and Haiming Wang and Hengzhi Wang and Huaqing Wang and Hui Wang and Jiahao Wang and Jinhong Wang and Jiuzheng Wang and Kaixin Wang and Linian Wang and Qibin Wang and Shengjie Wang and Shuyi Wang and Si Wang and Wei Wang and Xiaochen Wang and Xinyuan Wang and Yao Wang and Yejie Wang and Yipu Wang and Yiqin Wang and Yucheng Wang and Yuzhi Wang and Zhaoji Wang and Zhaowei Wang and Zhengtao Wang and Zhexu Wang and Zihan Wang and Zizhe Wang and Chu Wei and Ming Wei and Chuan Wen and Zichen Wen and Chengjie Wu and Haoning Wu and Junyan Wu and Rucong Wu and Wenhao Wu and Yuefeng Wu and Yuhao Wu and Yuxin Wu and Zijian Wu and Chenjun Xiao and Jin Xie and Xiaotong Xie and Yuchong Xie and Yifei Xin and Bowei Xing and Boyu Xu and Jianfan Xu and Jing Xu and Jinjing Xu and L. H. Xu and Lin Xu and Suting Xu and Weixin Xu and Xinbo Xu and Xinran Xu and Yangchuan Xu and Yichang Xu and Yuemeng Xu and Zelai Xu and Ziyao Xu and Junjie Yan and Yuzi Yan and Guangyao Yang and Hao Yang and Junwei Yang and Kai Yang and Ningyuan Yang and Ruihan Yang and Xiaofei Yang and Xinlong Yang and Ying Yang and Yi Yang and Yi Yang and Zhen Yang and Zhilin Yang and Zonghan Yang and Haotian Yao and Dan Ye and Wenjie Ye and Zhuorui Ye and Bohong Yin and Chengzhen Yu and Longhui Yu and Tao Yu and Tianxiang Yu and Enming Yuan and Mengjie Yuan and Xiaokun Yuan and Yang Yue and Weihao Zeng and Dunyuan Zha and Haobing Zhan and Dehao Zhang and Hao Zhang and Jin Zhang and Puqi Zhang and Qiao Zhang and Rui Zhang and Xiaobin Zhang and Y. Zhang and Yadong Zhang and Yangkun Zhang and Yichi Zhang and Yizhi Zhang and Yongting Zhang and Yu Zhang and Yushun Zhang and Yutao Zhang and Yutong Zhang and Zheng Zhang and Chenguang Zhao and Feifan Zhao and Jinxiang Zhao and Shuai Zhao and Xiangyu Zhao and Yikai Zhao and Zijia Zhao and Huabin Zheng and Ruihan Zheng and Shaojie Zheng and Tengyang Zheng and Junfeng Zhong and Longguang Zhong and Weiming Zhong and M. Zhou and Runjie Zhou and Xinyu Zhou and Zaida Zhou and Jinguo Zhu and Liya Zhu and Xinhao Zhu and Yuxuan Zhu and Zhen Zhu and Jingze Zhuang and Weiyu Zhuang and Ying Zou and Xinxing Zu},
      year={2026},
      eprint={2602.02276},
      archivePrefix={arXiv},
      primaryClass={cs.CL},
      url={https://arxiv.org/abs/2602.02276}, 
}

@article{cheng2025lmcache,
  title={LMCache: An Efficient KV Cache Layer for Enterprise-Scale LLM Inference},
  author={Cheng, Yihua and Liu, Yuhan and Yao, Jiayi and An, Yuwei and Chen, Xiaokun and Feng, Shaoting and Huang, Yuyang and Shen, Samuel and Du, Kuntai and Jiang, Junchen},
  journal={arXiv preprint arXiv:2510.09665},
  year={2025}
}

@misc{yang2025qwen3technicalreport,
      title={Qwen3 Technical Report}, 
      author={An Yang and Anfeng Li and Baosong Yang and Beichen Zhang and Binyuan Hui and Bo Zheng and Bowen Yu and Chang Gao and Chengen Huang and Chenxu Lv and Chujie Zheng and Dayiheng Liu and Fan Zhou and Fei Huang and Feng Hu and Hao Ge and Haoran Wei and Huan Lin and Jialong Tang and Jian Yang and Jianhong Tu and Jianwei Zhang and Jianxin Yang and Jiaxi Yang and Jing Zhou and Jingren Zhou and Junyang Lin and Kai Dang and Keqin Bao and Kexin Yang and Le Yu and Lianghao Deng and Mei Li and Mingfeng Xue and Mingze Li and Pei Zhang and Peng Wang and Qin Zhu and Rui Men and Ruize Gao and Shixuan Liu and Shuang Luo and Tianhao Li and Tianyi Tang and Wenbiao Yin and Xingzhang Ren and Xinyu Wang and Xinyu Zhang and Xuancheng Ren and Yang Fan and Yang Su and Yichang Zhang and Yinger Zhang and Yu Wan and Yuqiong Liu and Zekun Wang and Zeyu Cui and Zhenru Zhang and Zhipeng Zhou and Zihan Qiu},
      year={2025},
      eprint={2505.09388},
      archivePrefix={arXiv},
      primaryClass={cs.CL},
      url={https://arxiv.org/abs/2505.09388}, 
}

@misc{shoeybi2020megatronlmtrainingmultibillionparameter,
      title={Megatron-LM: Training Multi-Billion Parameter Language Models Using Model Parallelism}, 
      author={Mohammad Shoeybi and Mostofa Patwary and Raul Puri and Patrick LeGresley and Jared Casper and Bryan Catanzaro},
      year={2020},
      eprint={1909.08053},
      archivePrefix={arXiv},
      primaryClass={cs.CL},
      url={https://arxiv.org/abs/1909.08053}, 
}

@article{yao2023tree,
  title={Tree of thoughts: Deliberate problem solving with large language models},
  author={Yao, Shunyu and Yu, Dian and Zhao, Jeffrey and Shafran, Izhak and Griffiths, Tom and Cao, Yuan and Narasimhan, Karthik},
  journal={Advances in neural information processing systems},
  volume={36},
  pages={11809--11822},
  year={2023}
}

@article{xu2025mem,
  title={A-mem: Agentic memory for llm agents},
  author={Xu, Wujiang and Mei, Kai and Gao, Hang and Tan, Juntao and Liang, Zujie and Zhang, Yongfeng},
  journal={arXiv preprint arXiv:2502.12110},
  year={2025}
}

@misc{packer2024memgptllmsoperatingsystems,
      title={MemGPT: Towards LLMs as Operating Systems}, 
      author={Charles Packer and Sarah Wooders and Kevin Lin and Vivian Fang and Shishir G. Patil and Ion Stoica and Joseph E. Gonzalez},
      year={2024},
      eprint={2310.08560},
      archivePrefix={arXiv},
      primaryClass={cs.AI},
      url={https://arxiv.org/abs/2310.08560}, 
}

@misc{zhong2023memorybankenhancinglargelanguage,
      title={MemoryBank: Enhancing Large Language Models with Long-Term Memory}, 
      author={Wanjun Zhong and Lianghong Guo and Qiqi Gao and He Ye and Yanlin Wang},
      year={2023},
      eprint={2305.10250},
      archivePrefix={arXiv},
      primaryClass={cs.CL},
      url={https://arxiv.org/abs/2305.10250}, 
}

@misc{kim2024llmcompilerparallelfunction,
      title={An LLM Compiler for Parallel Function Calling}, 
      author={Sehoon Kim and Suhong Moon and Ryan Tabrizi and Nicholas Lee and Michael W. Mahoney and Kurt Keutzer and Amir Gholami},
      year={2024},
      eprint={2312.04511},
      archivePrefix={arXiv},
      primaryClass={cs.CL},
      url={https://arxiv.org/abs/2312.04511}, 
}

@article{zhang2024rest,
  title={Rest-mcts*: Llm self-training via process reward guided tree search},
  author={Zhang, Dan and Zhoubian, Sining and Hu, Ziniu and Yue, Yisong and Dong, Yuxiao and Tang, Jie},
  journal={Advances in Neural Information Processing Systems},
  volume={37},
  pages={64735--64772},
  year={2024}
}

@inproceedings{abdelaziz-etal-2024-granite,
    title = "Granite-Function Calling Model: Introducing Function Calling Abilities via Multi-task Learning of Granular Tasks",
    author = "Abdelaziz, Ibrahim  and
      Basu, Kinjal  and
      Agarwal, Mayank  and
      Kumaravel, Sadhana  and
      Stallone, Matthew  and
      Panda, Rameswar  and
      Rizk, Yara  and
      Bhargav, G P Shrivatsa  and
      Crouse, Maxwell  and
      Gunasekara, Chulaka  and
      Ikbal, Shajith  and
      Joshi, Sachindra  and
      Karanam, Hima  and
      Kumar, Vineet  and
      Munawar, Asim  and
      Neelam, Sumit  and
      Raghu, Dinesh  and
      Sharma, Udit  and
      Soria, Adriana Meza  and
      Sreedhar, Dheeraj  and
      Venkateswaran, Praveen  and
      Unuvar, Merve  and
      Cox, David Daniel  and
      Roukos, Salim  and
      Lastras, Luis A.  and
      Kapanipathi, Pavan",
    editor = "Dernoncourt, Franck  and
      Preo{\c{t}}iuc-Pietro, Daniel  and
      Shimorina, Anastasia",
    booktitle = "Proceedings of the 2024 Conference on Empirical Methods in Natural Language Processing: Industry Track",
    month = nov,
    year = "2024",
    address = "Miami, Florida, US",
    publisher = "Association for Computational Linguistics",
    url = "https://aclanthology.org/2024.emnlp-industry.85/",
    doi = "10.18653/v1/2024.emnlp-industry.85",
    pages = "1131--1139"
}

@misc{luo2025agentlightningtrainai,
      title={Agent Lightning: Train ANY AI Agents with Reinforcement Learning}, 
      author={Xufang Luo and Yuge Zhang and Zhiyuan He and Zilong Wang and Siyun Zhao and Dongsheng Li and Luna K. Qiu and Yuqing Yang},
      year={2025},
      eprint={2508.03680},
      archivePrefix={arXiv},
      primaryClass={cs.AI},
      url={https://arxiv.org/abs/2508.03680}, 
}

@misc{zhou2025sweetrltrainingmultiturnllm,
      title={SWEET-RL: Training Multi-Turn LLM Agents on Collaborative Reasoning Tasks}, 
      author={Yifei Zhou and Song Jiang and Yuandong Tian and Jason Weston and Sergey Levine and Sainbayar Sukhbaatar and Xian Li},
      year={2025},
      eprint={2503.15478},
      archivePrefix={arXiv},
      primaryClass={cs.LG},
      url={https://arxiv.org/abs/2503.15478}, 
}

@misc{zhang2025agentrlscalingagenticreinforcement,
      title={AgentRL: Scaling Agentic Reinforcement Learning with a Multi-Turn, Multi-Task Framework}, 
      author={Hanchen Zhang and Xiao Liu and Bowen Lv and Xueqiao Sun and Bohao Jing and Iat Long Iong and Zhenyu Hou and Zehan Qi and Hanyu Lai and Yifan Xu and Rui Lu and Hongning Wang and Jie Tang and Yuxiao Dong},
      year={2025},
      eprint={2510.04206},
      archivePrefix={arXiv},
      primaryClass={cs.AI},
      url={https://arxiv.org/abs/2510.04206}, 
}

@misc{shah2024flashattention3fastaccurateattention,
      title={FlashAttention-3: Fast and Accurate Attention with Asynchrony and Low-precision}, 
      author={Jay Shah and Ganesh Bikshandi and Ying Zhang and Vijay Thakkar and Pradeep Ramani and Tri Dao},
      year={2024},
      eprint={2407.08608},
      archivePrefix={arXiv},
      primaryClass={cs.LG},
      url={https://arxiv.org/abs/2407.08608}, 
}

@misc{wang2025flashmaskefficientrichmask,
      title={FlashMask: Efficient and Rich Mask Extension of FlashAttention}, 
      author={Guoxia Wang and Jinle Zeng and Xiyuan Xiao and Siming Wu and Jiabin Yang and Lujing Zheng and Zeyu Chen and Jiang Bian and Dianhai Yu and Haifeng Wang},
      year={2025},
      eprint={2410.01359},
      archivePrefix={arXiv},
      primaryClass={cs.LG},
      url={https://arxiv.org/abs/2410.01359}, 
}

@misc{qin2025mooncakekvcachecentricdisaggregatedarchitecture,
      title={Mooncake: A KVCache-centric Disaggregated Architecture for LLM Serving}, 
      author={Ruoyu Qin and Zheming Li and Weiran He and Mingxing Zhang and Yongwei Wu and Weimin Zheng and Xinran Xu},
      year={2025},
      eprint={2407.00079},
      archivePrefix={arXiv},
      primaryClass={cs.DC},
      url={https://arxiv.org/abs/2407.00079}, 
}

@misc{pope2022efficientlyscalingtransformerinference,
      title={Efficiently Scaling Transformer Inference}, 
      author={Reiner Pope and Sholto Douglas and Aakanksha Chowdhery and Jacob Devlin and James Bradbury and Anselm Levskaya and Jonathan Heek and Kefan Xiao and Shivani Agrawal and Jeff Dean},
      year={2022},
      eprint={2211.05102},
      archivePrefix={arXiv},
      primaryClass={cs.LG},
      url={https://arxiv.org/abs/2211.05102}, 
}

@misc{goru2025onepassreasontokenduplication,
      title={One-Pass to Reason: Token Duplication and Block-Sparse Mask for Efficient Fine-Tuning on Multi-Turn Reasoning}, 
      author={Ritesh Goru and Shanay Mehta and Prateek Jain},
      year={2025},
      eprint={2504.18246},
      archivePrefix={arXiv},
      primaryClass={cs.CL},
      url={https://arxiv.org/abs/2504.18246}, 
}

@article{Krell2021SequencePacking,
  title   = {Efficient Sequence Packing without Cross-contamination: Accelerating Large Language Models without Impacting Performance},
  author  = {Krell, Mario Michael and Kosec, Matej and Perez, Sergio P. and Fitzgibbon, Andrew},
  journal = {arXiv preprint arXiv:2107.02027},
  year    = {2021}
}

@misc{merrill2026terminalbenchbenchmarkingagentshard,
      title={Terminal-Bench: Benchmarking Agents on Hard, Realistic Tasks in Command Line Interfaces}, 
      author={Mike A. Merrill and Alexander G. Shaw and Nicholas Carlini and Boxuan Li and Harsh Raj and Ivan Bercovich and Lin Shi and Jeong Yeon Shin and Thomas Walshe and E. Kelly Buchanan and Junhong Shen and Guanghao Ye and Haowei Lin and Jason Poulos and Maoyu Wang and Marianna Nezhurina and Jenia Jitsev and Di Lu and Orfeas Menis Mastromichalakis and Zhiwei Xu and Zizhao Chen and Yue Liu and Robert Zhang and Leon Liangyu Chen and Anurag Kashyap and Jan-Lucas Uslu and Jeffrey Li and Jianbo Wu and Minghao Yan and Song Bian and Vedang Sharma and Ke Sun and Steven Dillmann and Akshay Anand and Andrew Lanpouthakoun and Bardia Koopah and Changran Hu and Etash Guha and Gabriel H. S. Dreiman and Jiacheng Zhu and Karl Krauth and Li Zhong and Niklas Muennighoff and Robert Amanfu and Shangyin Tan and Shreyas Pimpalgaonkar and Tushar Aggarwal and Xiangning Lin and Xin Lan and Xuandong Zhao and Yiqing Liang and Yuanli Wang and Zilong Wang and Changzhi Zhou and David Heineman and Hange Liu and Harsh Trivedi and John Yang and Junhong Lin and Manish Shetty and Michael Yang and Nabil Omi and Negin Raoof and Shanda Li and Terry Yue Zhuo and Wuwei Lin and Yiwei Dai and Yuxin Wang and Wenhao Chai and Shang Zhou and Dariush Wahdany and Ziyu She and Jiaming Hu and Zhikang Dong and Yuxuan Zhu and Sasha Cui and Ahson Saiyed and Arinbjörn Kolbeinsson and Jesse Hu and Christopher Michael Rytting and Ryan Marten and Yixin Wang and Alex Dimakis and Andy Konwinski and Ludwig Schmidt},
      year={2026},
      eprint={2601.11868},
      archivePrefix={arXiv},
      primaryClass={cs.SE},
      url={https://arxiv.org/abs/2601.11868}, 
}

@misc{cc,
  title        = {claude-code},
  url={https://claude.com/product/claude-code},
}

@misc{deepseekai2024deepseekv2strongeconomicalefficient,
      title={DeepSeek-V2: A Strong, Economical, and Efficient Mixture-of-Experts Language Model}, 
      author={DeepSeek-AI and Aixin Liu and Bei Feng and Bin Wang and Bingxuan Wang and Bo Liu and Chenggang Zhao and Chengqi Dengr and Chong Ruan and Damai Dai and Daya Guo and Dejian Yang and Deli Chen and Dongjie Ji and Erhang Li and Fangyun Lin and Fuli Luo and Guangbo Hao and Guanting Chen and Guowei Li and H. Zhang and Hanwei Xu and Hao Yang and Haowei Zhang and Honghui Ding and Huajian Xin and Huazuo Gao and Hui Li and Hui Qu and J. L. Cai and Jian Liang and Jianzhong Guo and Jiaqi Ni and Jiashi Li and Jin Chen and Jingyang Yuan and Junjie Qiu and Junxiao Song and Kai Dong and Kaige Gao and Kang Guan and Lean Wang and Lecong Zhang and Lei Xu and Leyi Xia and Liang Zhao and Liyue Zhang and Meng Li and Miaojun Wang and Mingchuan Zhang and Minghua Zhang and Minghui Tang and Mingming Li and Ning Tian and Panpan Huang and Peiyi Wang and Peng Zhang and Qihao Zhu and Qinyu Chen and Qiushi Du and R. J. Chen and R. L. Jin and Ruiqi Ge and Ruizhe Pan and Runxin Xu and Ruyi Chen and S. S. Li and Shanghao Lu and Shangyan Zhou and Shanhuang Chen and Shaoqing Wu and Shengfeng Ye and Shirong Ma and Shiyu Wang and Shuang Zhou and Shuiping Yu and Shunfeng Zhou and Size Zheng and T. Wang and Tian Pei and Tian Yuan and Tianyu Sun and W. L. Xiao and Wangding Zeng and Wei An and Wen Liu and Wenfeng Liang and Wenjun Gao and Wentao Zhang and X. Q. Li and Xiangyue Jin and Xianzu Wang and Xiao Bi and Xiaodong Liu and Xiaohan Wang and Xiaojin Shen and Xiaokang Chen and Xiaosha Chen and Xiaotao Nie and Xiaowen Sun and Xiaoxiang Wang and Xin Liu and Xin Xie and Xingkai Yu and Xinnan Song and Xinyi Zhou and Xinyu Yang and Xuan Lu and Xuecheng Su and Y. Wu and Y. K. Li and Y. X. Wei and Y. X. Zhu and Yanhong Xu and Yanping Huang and Yao Li and Yao Zhao and Yaofeng Sun and Yaohui Li and Yaohui Wang and Yi Zheng and Yichao Zhang and Yiliang Xiong and Yilong Zhao and Ying He and Ying Tang and Yishi Piao and Yixin Dong and Yixuan Tan and Yiyuan Liu and Yongji Wang and Yongqiang Guo and Yuchen Zhu and Yuduan Wang and Yuheng Zou and Yukun Zha and Yunxian Ma and Yuting Yan and Yuxiang You and Yuxuan Liu and Z. Z. Ren and Zehui Ren and Zhangli Sha and Zhe Fu and Zhen Huang and Zhen Zhang and Zhenda Xie and Zhewen Hao and Zhihong Shao and Zhiniu Wen and Zhipeng Xu and Zhongyu Zhang and Zhuoshu Li and Zihan Wang and Zihui Gu and Zilin Li and Ziwei Xie},
      year={2024},
      eprint={2405.04434},
      archivePrefix={arXiv},
      primaryClass={cs.CL},
      url={https://arxiv.org/abs/2405.04434}, 
}

@misc{yang2025swesmithscalingdatasoftware,
      title={SWE-smith: Scaling Data for Software Engineering Agents}, 
      author={John Yang and Kilian Lieret and Carlos E. Jimenez and Alexander Wettig and Kabir Khandpur and Yanzhe Zhang and Binyuan Hui and Ofir Press and Ludwig Schmidt and Diyi Yang},
      year={2025},
      eprint={2504.21798},
      archivePrefix={arXiv},
      primaryClass={cs.SE},
      url={https://arxiv.org/abs/2504.21798}, 
}

@misc{su2023roformerenhancedtransformerrotary,
      title={RoFormer: Enhanced Transformer with Rotary Position Embedding},
      author={Jianlin Su and Yu Lu and Shengfeng Pan and Ahmed Murtadha and Bo Wen and Yunfeng Liu},
      year={2023},
      eprint={2104.09864},
      archivePrefix={arXiv},
      primaryClass={cs.CL},
      url={https://arxiv.org/abs/2104.09864},
}

@misc{yang2025gateddeltanetworksimproving,
      title={Gated Delta Networks: Improving Mamba2 with Delta Rule}, 
      author={Songlin Yang and Jan Kautz and Ali Hatamizadeh},
      year={2025},
      eprint={2412.06464},
      archivePrefix={arXiv},
      primaryClass={cs.CL},
      url={https://arxiv.org/abs/2412.06464}, 
}

@misc{dao2024transformersssmsgeneralizedmodels,
      title={Transformers are SSMs: Generalized Models and Efficient Algorithms Through Structured State Space Duality}, 
      author={Tri Dao and Albert Gu},
      year={2024},
      eprint={2405.21060},
      archivePrefix={arXiv},
      primaryClass={cs.LG},
      url={https://arxiv.org/abs/2405.21060}, 
}

@misc{gu2024mambalineartimesequencemodeling,
      title={Mamba: Linear-Time Sequence Modeling with Selective State Spaces}, 
      author={Albert Gu and Tri Dao},
      year={2024},
      eprint={2312.00752},
      archivePrefix={arXiv},
      primaryClass={cs.LG},
      url={https://arxiv.org/abs/2312.00752}, 
}

@misc{qwen35blog,
    title = {Qwen3.5: Accelerating Productivity with Native Multimodal Agents},
    url = {https://qwen.ai/blog?id=qwen3.5},
    author = {Qwen Team},
    month = {February},
    year = {2026}
}

@software{ortools,
  title = {OR-Tools},
  version = {v9.10},
  author = {Laurent Perron and Vincent Furnon},
  organization = {Google},
  url = {https://developers.google.com/optimization/},
  date = {2024-05-07}
}

@misc{gupta2026persistent,
      title={A Persistent-State Dataflow Accelerator for Memory-Bound Linear Attention Decode on {FPGA}},
      author={Neelesh Gupta and Peter Wang and Rajgopal Kannan and Viktor K. Prasanna},
      year={2026},
      eprint={2603.05931},
      archivePrefix={arXiv},
      primaryClass={cs.AR},
      url={https://arxiv.org/abs/2603.05931},
}
\bibliographystyle{wwjjhh2025}

%%%%%%%%%%%%%%%%%%%%%%%%%%%%%%%%%%%%%%%%%%%%%%%%%%%%%%%%%%%%%%%%%%%%%%%%%%%%%%%
%%%%%%%%%%%%%%%%%%%%%%%%%%%%%%%%%%%%%%%%%%%%%%%%%%%%%%%%%%%%%%%%%%%%%%%%%%%%%%%
% SUPPLEMENTAL CONTENT AS APPENDIX AFTER REFERENCES
%%%%%%%%%%%%%%%%%%%%%%%%%%%%%%%%%%%%%%%%%%%%%%%%%%%%%%%%%%%%%%%%%%%%%%%%%%%%%%%
%%%%%%%%%%%%%%%%%%%%%%%%%%%%%%%%%%%%%%%%%%%%%%%%%%%%%%%%%%%%%%%%%%%%%%%%%%%%%%%
\clearpage
\onecolumn
\appendix

%% ============================================================
%% Appendix A: DFS Serialization — Implementation Details
%% ============================================================

\section{DFS Serialization: Implementation Details}
\label{app:dfs_impl}

\subsection{Tree Attention Mask via FlashMask}

The tree attention mask allows token $i$ to attend to token $j$ iff $j$ is in an
ancestor chunk of $i$ (full access to the entire ancestor chunk) or in the same chunk
with $j \le i$ (causal within chunk).
We implement this as a block-sparse mask using
FlashMask~\cite{wang2025flashmaskefficientrichmask}, which extends FlashAttention to
support rich per-query start/end column bounds and skips masked blocks entirely:

\begin{lstlisting}[language=Python]
attn_mask_bool = torch.zeros(N, N, dtype=torch.bool)
for c in range(n_chunks):
    c_start, c_end = chunk_boundaries[c]
    for anc in ancestors[c]:          # ancestors[c] includes c itself
        a_start, a_end = chunk_boundaries[anc]
        for i in range(c_start, c_end):
            if anc == c:              # causal within own chunk
                attn_mask_bool[i, c_start : i + 1] = True
            else:                     # full access to ancestor chunk
                attn_mask_bool[i, a_start : a_end] = True
\end{lstlisting}

\subsection{Tree-Aware SSM State Routing}

Standard chunk GatedDeltaNet passes the recurrent state sequentially: chunk $i$ reads
the output state of chunk $i{-}1$.
Under DFS serialization, after a leaf chunk the sequence backtracks to a sibling branch,
so the sequential predecessor is a sibling, not an ancestor.
We replace sequential routing with tree routing: each chunk reads the output state of its
\emph{parent} chunk.
\texttt{all\_states[0]} holds the zero initial state;
\texttt{all\_states[i+1]} holds the state produced after processing chunk $i$.
The index arithmetic \texttt{parent\_list\_idx = chunk\_parent\_map[i] + 1} maps root
chunks (\texttt{parent\_map}$=-1$) to index~0 and non-root chunks to their parent's
output state.
DFS pre-order guarantees the parent state is always computed before any child requests it.
Sibling chunks reference the same parent state tensor; PyTorch autograd accumulates their
gradient contributions there automatically.

\begin{lstlisting}[language=Python]
def torch_chunk_gated_delta_rule_tree_varlen(
    query, key, value, g, beta,
    chunk_boundaries,   # list[(start, end)] token index range per tree node
    chunk_parent_map,   # list[int] parent chunk idx; -1 = use initial_state
    initial_state=None,
):
    init = (initial_state.float() if initial_state is not None
            else torch.zeros(B, H, Dk, Dv, dtype=torch.float32, device=dev))

    all_states  = [init]   # all_states[0]=init, all_states[i+1]=state after chunk i
    all_outputs = []

    for i, (c_start, c_end) in enumerate(chunk_boundaries):
        L = c_end - c_start
        q_i, k_i, v_i = q[:,:,c_start:c_end], k[:,:,c_start:c_end], v[:,:,c_start:c_end]
        bt_i, g_i     = bt[:,:,c_start:c_end], g[:,:,c_start:c_end]

        v_beta_i = v_i * bt_i.unsqueeze(-1)
        k_beta_i = k_i * bt_i.unsqueeze(-1)
        g_cum    = g_i.cumsum(dim=-1)                              # [B, H, L]
        decay_mask = (g_cum.unsqueeze(-1) - g_cum.unsqueeze(-2)).tril().exp().tril()

        # ---- Within-chunk SSM correction ----
        attn_raw = -(k_beta_i @ k_i.transpose(-1,-2) * decay_mask).masked_fill(triu_mask, 0)
        attn_rows = [attn_raw[..., 0:1, :]]
        for j in range(1, L):
            attn_rows.append(attn_raw[..., j:j+1, :] + attn_raw[..., j:j+1, :j] @ attn_raw[..., :j, :])
        attn       = torch.cat(attn_rows, dim=-2) + torch.eye(L, device=dev)
        value_corr = attn @ v_beta_i                               # [B, H, L, Dv]
        k_cumdecay = attn @ (k_beta_i * g_cum.exp().unsqueeze(-1)) # [B, H, L, Dk]

        # ---- Cross-chunk: read from PARENT state (tree routing) ----
        parent_list_idx = int(chunk_parent_map[i]) + 1  # -1->0, j->j+1
        current_state   = all_states[parent_list_idx]   # [B, H, Dk, Dv]

        v_prime     = k_cumdecay @ current_state         # [B, H, L, Dv]
        v_new       = value_corr - v_prime
        attn_within = (q_i @ k_i.transpose(-1,-2) * decay_mask).masked_fill(triu_excl, 0)
        attn_inter  = (q_i * g_cum.exp().unsqueeze(-1)) @ current_state
        out_i       = attn_inter + attn_within @ v_new   # [B, H, L, Dv]
        all_outputs.append(out_i)

        # ---- Store new state for children ----
        new_state = (
            current_state * g_cum[:, :, -1, None, None].exp()
            + (k_i * (g_cum[:, :, -1:] - g_cum).exp().unsqueeze(-1)).transpose(-1,-2) @ v_new
        )
        all_states.append(new_state)   # sibling chunks will both read current_state above

    core_out = torch.cat(all_outputs, dim=2).transpose(1, 2)  # [B, S, H, Dv]
    return core_out
\end{lstlisting}

\subsection{Tree-Correct Causal Convolution}

GDN layers include a causal conv1d with kernel size $K_{\mathrm{conv}}$.
At a DFS backtrack position, the $K_{\mathrm{conv}}{-}1$ tokens immediately before a
child chunk belong to a sibling branch, not the ancestor path, so using them as context
is incorrect.
We fix this by having each chunk prepend its \emph{parent}'s saved last
$K_{\mathrm{conv}}{-}1$ effective tokens as left context, then discard them from the
output.
The saved context is taken from \texttt{ext[:, :, -(K-1):]}, i.e.\ the tail of the
concatenated \texttt{[parent\_ctx, chunk]} tensor---which includes ancestor history---so
grandchildren see the full $K_{\mathrm{conv}}{-}1$ ancestor token history, matching an
independent per-path forward.

\begin{lstlisting}[language=Python]
def _tree_correct_conv_varlen(ssm, mixed_qkv_raw, chunk_boundaries, chunk_parent_map):
    K             = ssm.conv_kernel_size
    chunk_outputs = []
    conv_states   = {}   # chunk_idx -> [B, conv_dim, K-1] saved context for children

    for c, (c_start, c_end) in enumerate(chunk_boundaries):
        L     = c_end - c_start
        chunk = mixed_qkv_raw[:, :, c_start:c_end]         # [B, conv_dim, L]

        parent_state = conv_states.get(chunk_parent_map[c], None)

        if parent_state is not None:
            ext       = torch.cat([parent_state, chunk], dim=-1)  # [B, conv_dim, K-1+L]
            out       = ssm.causal_conv1d_fn(
                            x=ext, weight=ssm.conv1d.weight.squeeze(1),
                            bias=ssm.conv1d.bias, activation=ssm.activation)
            chunk_out = out[:, :, K - 1:]    # drop prepended context -> [B, conv_dim, L]
        else:
            chunk_out = ssm.causal_conv1d_fn(
                            x=chunk, weight=ssm.conv1d.weight.squeeze(1),
                            bias=ssm.conv1d.bias, activation=ssm.activation)

        chunk_outputs.append(chunk_out)

        # Save last K-1 EFFECTIVE tokens (ancestor ctx + chunk tail) for children.
        # Using ext tail (not just chunk) propagates ancestor history to grandchildren.
        if parent_state is not None:
            conv_states[c] = ext[:, :, -(K - 1):].clone()
        else:
            conv_states[c] = F.pad(chunk, (max(0, K-1-L), 0))[:, :, -(K-1):].clone()

    return torch.cat(chunk_outputs, dim=2)   # [B, conv_dim, N]
\end{lstlisting}

%% ============================================================
%% FILE: appendix_hybrid.tex
%%
%% Add to appendix in main.tex:
%%   \input{appendix_hybrid.tex}
%% ============================================================

\section{Zero-Overhead Partition Boundaries: Implementation Details}
\label{app:hybrid_impl}

This appendix provides the complete implementation details for the KV-cache
gradient checkpointing at partition boundaries described in
Section~\ref{tpack}.

\subsection{Data Structures}

\paragraph{SubtreeGateway.}
Each cut node~$n_c$ in partition~$P$ produces one gateway for the child partition:

\begin{itemize}
  \item $\{(\tilde{\mathbf{G}}^{(l)}_{n_c}, \tilde{\mathbf{V}}^{(l)}_{n_c})\}_l$:
    detached leaf KV tensors with \texttt{requires\_grad=True},
    shape $[1, H, e_{n_c}, d]$ per layer.
  \item $\{(\mathbf{G}^{(l)}_{n_c}, \mathbf{V}^{(l)}_{n_c})\}_l$:
    original KV slices retaining \texttt{grad\_fn}.
  \item Float32 accumulator hooks: registered on each leaf tensor to capture
    gradient contributions in float32 before bfloat16 rounding; see below.
  \item $\mathtt{attn\_bias} \in \mathbb{R}^{e_{n_c}}$: additive attention
    bias with $0$ at ancestor positions and $-\infty$ at non-ancestor positions
    (Section~\ref{app:anc_mask}).
  \item $\mathtt{pos\_offset}$: depth-based position of the child's first token
    (Section~\ref{app:pos_offset}).
  \item For SSM hybrid: $\{(\tilde{\mathbf{S}}^{(l)}_{n_c}, \tilde{\mathbf{C}}^{(l)}_{n_c})\}_l$
    and their raw counterparts (Section~\ref{app:ssm_gw}).
\end{itemize}

\subsection{Forward Pass}

\paragraph{Parent partition forward.}
Run DFS forward on $P$'s token sequence with the standard tree attention mask
and SSM patch (Section~\ref{impl}).  The model is called with
$\texttt{past\_key\_values} = \text{GatewayCache}(\tilde{\mathbf{G}}_{\mathrm{parent}})$
if $P$ itself is a child partition, otherwise an empty cache.

After the forward, for each cut node~$n_c \in P$:
\begin{enumerate}
  \item Slice the accumulated KV:
    $\mathbf{G}^{(l)}_{n_c} \leftarrow \mathbf{K}^{(l)}_{[\,:\,\mathtt{past\_len}+e_{n_c}]}$,
    which retains \texttt{grad\_fn} via the concatenation inside GatewayCache.
  \item Detach to create gateway leaf tensors and register float32 accumulator hooks:
  \begin{equation}
    \tilde{\mathbf{G}}^{(l)}_{n_c} = \mathrm{detach}\!\left(\mathbf{G}^{(l)}_{n_c}\right).\,\mathrm{requires\_grad}(\mathtt{True}).
    \label{eq:gateway}
  \end{equation}
\end{enumerate}

\paragraph{Child partition forward.}
Run DFS forward on $C$'s own token sequence with the \emph{combined} attention mask:
\begin{equation}
  M_{ij} = \begin{cases}
    \mathtt{attn\_bias}_j & j < \mathtt{past\_len} \\
    m^{\mathrm{tree}}_{i,\,j-\mathtt{past\_len}} & j \geq \mathtt{past\_len}
  \end{cases}
  \label{eq:combined_mask}
\end{equation}
where $m^{\mathrm{tree}}$ is the standard within-subtree tree attention mask.
The model receives $\texttt{past\_key\_values} = \text{GatewayCache}(\tilde{\mathbf{G}}_{n_c})$.

\subsection{Ancestor-Aware Attention Bias}
\label{app:anc_mask}

In a DFS-ordered partition, the gateway slice $[\,:\,e_{n_c}]$ may include tokens
from sibling subtrees that are \emph{not} ancestors of~$n_c$.
For example, if $P$'s DFS sequence is $[a, b, c_1, c_2]$ where $c_1, c_2$ are siblings
and $n_c = c_2$, the gateway includes $c_1$'s tokens, which $n_c$'s children must not attend to.
We compute ancestor positions by traversing $n_c$'s parent chain within~$P$:
\begin{equation}
  \mathtt{anc}(n_c, P) = \bigcup_{n' \in \mathrm{path}(P.\mathrm{root}, n_c)} [s_{n'}, e_{n'}]
\end{equation}
where $[s_{n'}, e_{n'}]$ is the token range of $n'$ in $P$'s DFS sequence.
The attention bias is:
\begin{equation}
  \mathtt{attn\_bias}_j =
  \begin{cases}
    0 & j \in \mathtt{anc}(n_c, P) \cup [0, \mathtt{past\_len}_P) \\
    -\infty & \text{otherwise}
  \end{cases}
\end{equation}
(The term $[0, \mathtt{past\_len}_P)$ covers $P$'s own gateway past, which is always
fully visible since it was already ancestor-filtered when~$P$'s gateway was created.)

\subsection{Depth-Based Position Offset}
\label{app:pos_offset}

The child partition's position IDs must match those that $n_c$'s children would
receive in a standalone per-path forward (to ensure RoPE produces identical results).
Let $\mathrm{pos}^{(P)}_{n_c}$ be the depth-based position of $n_c$'s \emph{first} token
within~$P$ (obtained from $P$'s \texttt{dfs\_pos\_ids} array).  Then:
\begin{equation}
  \mathtt{pos\_offset}(C) = \mathtt{pos\_offset}(P) + \mathrm{pos}^{(P)}_{n_c} + |T_{n_c}|
  \label{eq:pos_offset_child}
\end{equation}
Critically, $\mathtt{pos\_offset}(C) \neq \mathtt{past\_len}$ when $P$'s DFS contains
sibling tokens between $P$'s first token and~$n_c$: $\mathtt{past\_len}$ counts all
gateway tokens (including non-ancestor siblings), whereas~\eqref{eq:pos_offset_child}
counts only ancestor tokens up to and including~$n_c$.

\subsection{Float32 Gradient Accumulation}
\label{app:fp32_hook}

When multiple child partitions share the same cut node (i.e., $n_c$ has
$k > 1$ children going to distinct child partitions), all children's backward
passes accumulate gradients into the same leaf gateway tensors.
Direct accumulation into the bfloat16 \texttt{.grad} attribute introduces rounding
errors from repeated bfloat16 additions.
We register a float32 accumulator hook on each leaf tensor:
\begin{equation}
  \tilde{\mathbf{G}}^{(l)}.\texttt{register\_hook}\!\left(\lambda g: \mathrm{acc}^{(l)} \mathrel{+}= g.\mathrm{float}(),\; \mathrm{return}\; \mathbf{0}\right)
\end{equation}
Returning $\mathbf{0}$ suppresses writes to the bfloat16 \texttt{.grad}, so the float32
accumulator captures the exact sum.  The chain backward~\eqref{eq:gateway_chain}
then uses the float32 accumulated values.
Note: even though we pass float32 gradients to \texttt{autograd.backward},
PyTorch's type-promotion rules cast them back to bfloat16 when entering a bfloat16
matmul; the hook therefore primarily eliminates \emph{inter-child accumulation} rounding,
not the in-kernel computation.

\subsection{Backward Pass and retain\_graph Scheduling}

Processing partition~$P$ in reverse topological order:
\begin{enumerate}
  \item For each actual leaf $\ell \in P$: compute $\mathcal{L}_\ell$ and call
    $\mathcal{L}_\ell.\mathtt{backward}(\mathtt{retain\_graph}=\mathtt{True})$.
  \item For each unique cut node~$n_c$ of~$P$ (deduplicated to avoid double-counting
    when $n_c$ has multiple children): chain gradients back via:
    \begin{equation}
      \texttt{autograd.backward}\!\left(\!\left\{\mathbf{G}^{(l)}_{n_c}\right\}_l,\;
      \left\{\nabla_{\tilde{\mathbf{G}}^{(l)}_{n_c}}\right\}_l\!\right)
      \label{eq:gateway_chain}
    \end{equation}
    with $\mathtt{retain\_graph}=\mathtt{True}$ for all but the last operation,
    $\mathtt{False}$ for the last.
\end{enumerate}
Total backward traversals of $P$'s computation graph
$= |\{\ell \in P : \ell \text{ is a leaf}\}| + |\text{unique cut nodes of } P|$.
The graph is freed after the last traversal.

\subsection{SSM State Gateway}
\label{app:ssm_gw}

\paragraph{State extraction.}
The existing DFS SSM forward (\texttt{torch\_chunk\_gated\_delta\_rule\_tree\_varlen})
maintains an \texttt{all\_states} list where \texttt{all\_states[i+1]} is the recurrent
state after chunk~$i$.
We extend it with a \texttt{capture\_states\_at} argument: after computing
\texttt{new\_state} for chunk~$c$, if $c \in \mathtt{capture\_states\_at}$,
store \texttt{all\_states[c+1]} (which has \texttt{grad\_fn}).
Similarly, \texttt{\_tree\_correct\_conv\_varlen} is extended to return
\texttt{conv\_states[c]} (the $K{-}1$ conv context after chunk~$c$) at requested positions.

\paragraph{State injection.}
For the child partition's SSM forward, we inject gateway states via two mechanisms:
\begin{itemize}
  \item \textit{Recurrent state}: pass $\tilde{\mathbf{S}}^{(l)}_{n_c}$ as
    \texttt{initial\_state} to \texttt{torch\_chunk\_gated\_delta\_rule\_tree\_varlen}.
    Root chunks (those with \texttt{parent\_map}\,$=-1$) read from
    \texttt{all\_states[0]} $=$ \texttt{initial\_state}, matching the gateway.
  \item \textit{Conv context}: pre-populate \texttt{conv\_states[-1]}
    $= \tilde{\mathbf{C}}^{(l)}_{n_c}$ before calling \texttt{\_tree\_correct\_conv\_varlen}.
    Root chunks look up \texttt{conv\_states.get(-1, None)} as their parent context,
    finding the gateway value.
\end{itemize}
This design requires no structural changes to the SSM forward; the injection is
entirely through the existing lookup mechanism.

\paragraph{Gradient chaining.}
The raw tensors $\mathbf{S}^{(l)}_{n_c}$ and $\mathbf{C}^{(l)}_{n_c}$ (with \texttt{grad\_fn})
are included in the \texttt{autograd.backward} call alongside the KV tensors.
All KV and SSM gradients are chained in a single call per cut node, with a shared
\texttt{retain\_graph} flag.

\subsection{Numerical Verification}

\paragraph{Pure-attention models (Qwen3-4B).}
\begin{itemize}
  \item \emph{Self-consistency}: two runs with the same inputs give EXACT 0.
  \item \emph{Backward vs.\ reference (float32)}: max-relative error $< 10^{-4}$
    across max-tokens $\in \{1000, 20, 10\}$ (verified against \texttt{tree\_grad\_ckpt\_sequential}).
\end{itemize}

\paragraph{SSM hybrid models (Qwen3.5, 2-layer float32).}
\begin{itemize}
  \item \emph{Self-consistency}: EXACT 0 across all max-tokens settings.
  \item \emph{Backward vs.\ reference (float32)}: max-relative error $< 2 \times 10^{-5}$,
    consistent across no-cut (max-tokens\,=\,1000) and aggressively-cut
    (max-tokens\,=\,3) settings, confirming that the SSM state gateway introduces
    no additional numerical error beyond standard float32 arithmetic.
  \item \emph{Forward}: differences of $\sim 5 \times 10^{-6}$ between partition sizes
    arise from different DFS sequence lengths feeding into float32 operations;
    two runs with identical partition structure give EXACT 0.
\end{itemize}

\end{document}